%% file: acl.tex
\pdfoutput=1

\documentclass[11pt]{article}

\usepackage[]{acl}
\usepackage{bm}

\usepackage{times}
\usepackage{latexsym}

\usepackage[T1]{fontenc}

\usepackage[utf8]{inputenc}

\usepackage{microtype}

\usepackage{amsfonts}
\usepackage{amsmath}
\usepackage{amssymb}
\usepackage{subcaption}
\usepackage{caption}
\usepackage{multicol}
\usepackage{multirow}
\usepackage{booktabs}
\usepackage{graphicx}
\usepackage{float}
\usepackage{efbox}
\usepackage{subcaption}
\usepackage{cleveref}
\usepackage{url}
\usepackage{amssymb}
\usepackage{pifont}
\usepackage[ruled, lined, linesnumbered, commentsnumbered, longend]{algorithm2e}
\def\@fnsymbol#1{\ensuremath{\ifcase#1\or *\or \dagger\or \ddagger\or
   \mathsection\or \mathparagraph\or \|\or **\or \dagger\dagger
   \or \ddagger\ddagger \else\@ctrerr\fi}}
\newcommand{\ssymbol}[1]{^{\@fnsymbol{#1}}}

\title{New Intent Discovery with Pre-training and Contrastive Learning}

\author{
Yuwei Zhang$^2$\thanks{~ Work done while the author was with HK PolyU.} \quad Haode Zhang$^1$ \quad Li-Ming Zhan$^{1}$ \quad \quad Albert Y.S. Lam$^3$ \\
\bf Xiao-Ming Wu$^1$\Thanks{~ Corresponding author.} \\ 
Department of Computing, The Hong Kong Polytechnic University, Hong Kong S.A.R.$^1$ \\
University of California, San Diego$^2$\\
Fano Labs, Hong Kong S.A.R.$^3$ \\
{\tt zhangyuwei.work@gmail.com}\\
{\tt \{haode.zhang, lmzhan.zhan\}@connect.polyu.edu.hk}\\
{\tt csxmwu@comp.polyu.edu.hk, albert@fano.ai}\\
}

\begin{document}
\maketitle
\begin{abstract}
New intent discovery aims to uncover novel intent categories from user utterances to expand the set of supported intent classes. It is a critical task for the development and service expansion of a practical dialogue system. Despite its importance, this problem remains under-explored in the literature. Existing approaches typically rely on a large amount of labeled utterances and employ pseudo-labeling methods for representation learning and clustering, which are label-intensive, inefficient, and inaccurate. 
In this paper, we provide new solutions to two important research questions for new intent discovery: (1) how to learn semantic utterance representations and (2) how to better cluster utterances. Particularly, we first propose a multi-task pre-training strategy to leverage rich unlabeled data along with external labeled data for representation learning.
Then, we design a new contrastive loss to exploit self-supervisory signals in unlabeled data for clustering.
Extensive experiments on three intent recognition benchmarks demonstrate the high effectiveness of our proposed method, which outperforms state-of-the-art methods by a large margin in both unsupervised and semi-supervised scenarios. The source code will be available at \url{https://github.com/zhang-yu-wei/MTP-CLNN}.
\end{abstract}

\input{sections/introduction}
\input{sections/related}
\input{sections/method}
\input{sections/experiment}
\input{sections/conclusion}
\clearpage
\bibliography{anthology,custom}
\bibliographystyle{acl_natbib}

\clearpage
\input{appendix}
\end{document}

%% file: sections/introduction.tex
\section{Introduction}
\label{intro}

\begin{figure}[t]
    \centering
    \includegraphics[width=7.7cm]{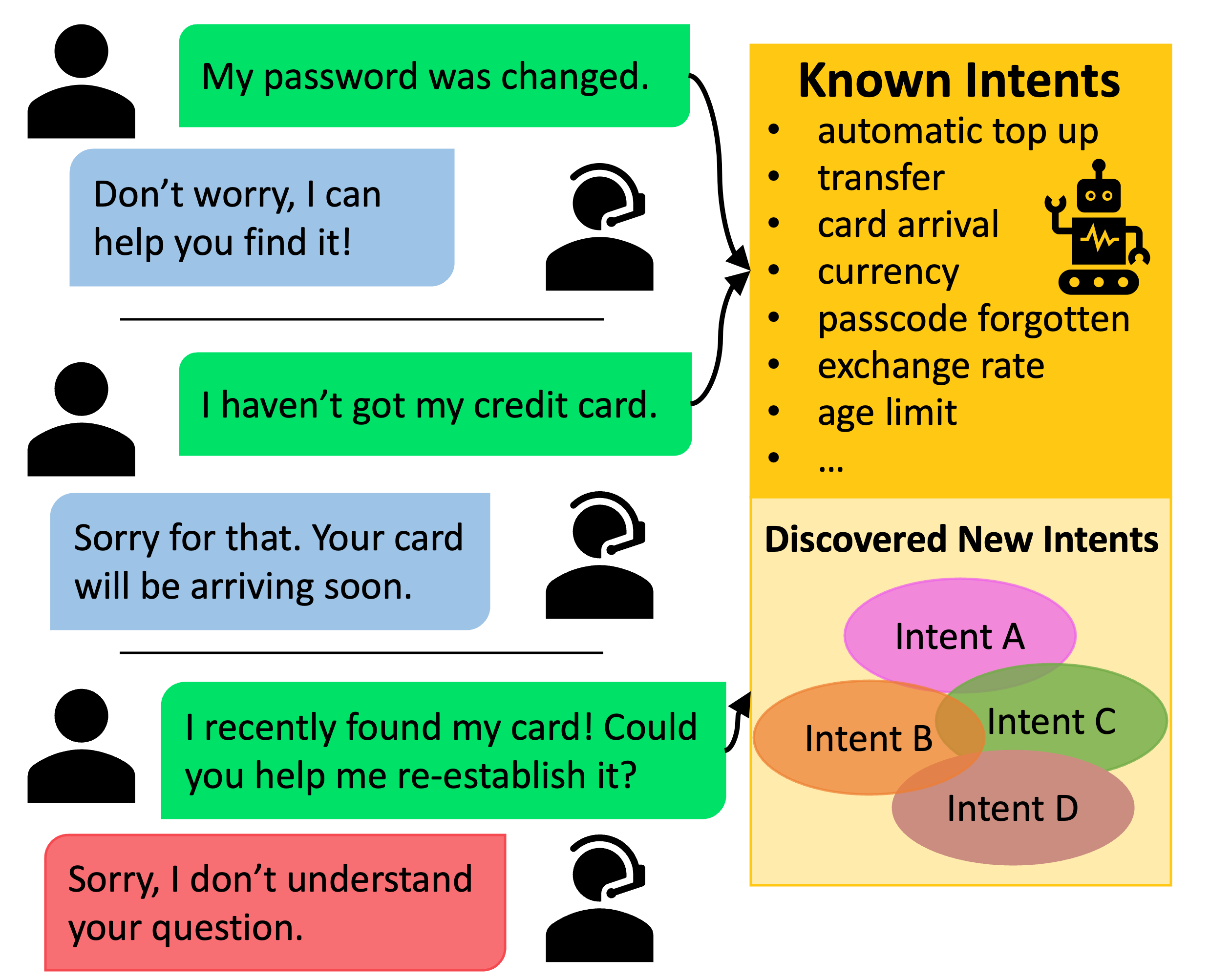}
    \caption{New Intent Discovery.}
    \label{fig:nid}
\end{figure}

\textbf{Why Study New Intent Discovery (NID)?}
Recent years have witnessed the rapid growth of conversational AI applications.
To design a natural language understanding system, a set of expected customer intentions are collected 
beforehand to train an intent recognition model.
However, the pre-defined intents 
cannot fully meet customer needs.
This implies the necessity 
of expanding the intent recognition model
by repeatedly integrating new intents discovered from unlabeled user utterances (Fig.~\ref{fig:nid}).
To reduce the effort in manually identifying unknown intents from a mass of utterances, previous works commonly employ clustering algorithms to group utterances of similar intents~\citep{cheung2012sequence,hakkani2015clustering,padmasundari2018intent}.
The cluster assignments thereafter can either be directly used as new intent labels or as heuristics for faster annotations.

\textbf{Research Questions (RQ) and Challenges.} Current study of NID centers around two basic research questions: \emph{1) How to learn semantic utterance representations to provide proper cues for clustering? 2) How to better cluster the utterances? } The study of the two questions are often interwoven in existing research.
Utterances can be represented according to different aspects such as the style of language, the related topics, or even the length of sentences. It is important to learn semantic utterance representations to provide proper cues for clustering. Simply applying a vanilla pre-trained language model (PLM) to generate utterance representations is not a viable solution, which leads to poor performance on NID as shown by the experimental results in Section~\ref{exp:result}.
Some recent works proposed to use labeled utterances of known intents for representation learning~\citep{forman2015semi,haponchyk2018structured,lin2020constrained,zhang2021aligned,haponchyk2021structured},
but they require a substantial amount of known intents and sufficient labeled utterances of each intent, which are not always available especially at the early development stage of a dialogue system. 
Further, pseudo-labeling approaches are often exploited to generate supervision signals for representation learning and clustering. For example, \citet{lin2020constrained} finetune a PLM with an utterance similarity prediction task on labeled utterances to guide the training of unlabeled data with pseudo-labels. \citet{zhang2021aligned} adopt a deep clustering method \citep{caron2018deep} that uses $k$-means clustering to produce pseudo-labels. 
However, pseudo-labels are often noisy and can lead to error propagation. 
\textbf{Our Solutions.}
In this work, we propose a simple yet effective solution for each research question.
\textbf{Solution to RQ 1: multi-task pre-training.} 
We propose a multi-task pre-training strategy that takes advantage of both external data and internal data for representation learning. Specifically, we leverage publicly available, high-quality intent detection datasets, following \citet{zhang2021effectiveness}, as well as the provided labeled and unlabeled utterances in the current domain, to fine-tune a pre-trained PLM to learn task-specific utterance representations for NID. The multi-task learning strategy enables knowledge transfer from general intent detection tasks and adaptation to a specific application domain.
\textbf{Solution to RQ 2: contrastive learning with nearest neighbors.} 
We propose to use a contrastive loss to produce compact clusters, which is motivated by the recent success of contrastive learning in both computer vision \citep{bachman2019cpc,he2019moco,chen2020simple,khosla2020supervised} and natural language processing \citep{gunel2021supervised,gao2021simcse,yan2021consert}.
Contrastive learning usually maximizes the agreement between different views of the same example and minimize that between different examples.
However, the commonly used instance discrimination task may push away false negatives and hurts the clustering performance.
Inspired by a recent work in computer vision~\citep{vangansbeke2020scan}, 
we introduce neighborhood relationship to customize the contrastive loss for clustering in both unsupervised (i.e., without any labeled utterances of known intents) and semi-supervised scenarios.
Intuitively, in a semantic feature space, neighboring utterances should have a similar intent, and pulling together neighboring samples makes clusters more compact.

Our main contributions are three-fold.
\begin{itemize}
    
    \item We show that our proposed multi-task pre-training method already leads to large performance gains over state-of-the-art models for both unsupervised and semi-supervised NID. 
    
    \item We propose a self-supervised clustering method for NID by incorporating neighborhood relationship into the contrastive learning objective, which further boosts performance.
    
    \item We conduct extensive experiments and ablation studies on three benchmark datasets to verify the effectiveness of our methods.
    
\end{itemize}

%% file: sections/related.tex
\section{Related Works}

\textbf{New Intent Discovery.}
The study of NID is still in an early stage.
Pioneering works focus on unsupervised clustering methods. \citet{shi2018auto} leveraged auto-encoder to extract features.
\citet{perkins2019multiview} considered the context of an utterance in a conversation.
\citet{chatterjee2020mining} proposed to improve density-based models.
Some recent works~\cite{haponchyk2018structured,haponchyk2021structured} studied supervised clustering algorithms for intent labeling, yet it can not handle new intents.
Another line of works~\cite{forman2015semi,lin2020constrained,zhang2021aligned} investigated a more practical case where some known intents are provided to support the discovery of unknown intents, which is often referred to as semi-supervised NID.

To tackle semi-supervised NID, \citet{lin2020constrained} proposed to first perform supervised training on known intents with a sentence similarity task and then use pseudo labeling on unlabeled utterances to learn a better embedding space.
\citet{zhang2021aligned} proposed to first pre-train on known intents and then perform $k$-means clustering to assign pseudo labels on unlabeled data for representation learning following Deep Clustering~\citep{caron2018deep}. They also proposed to align clusters to accelerate the learning of top layers.
Another approach is to first classify the utterances as known and unknown and then uncover new intents with the unknown utterances \citep{vedula2020automatic,zhang2021textoir}. Hence, it relies on accurate classification in the first stage.

In this work, we address NID by proposing a multi-task pre-training method for representation learning and a contrastive learning method for clustering. In contrast to previous methods that rely on ample annotated data in the current domain for pre-training, our method can be used in an unsupervised setting and work well in data-scarce scenarios (Section~\ref{exp:pki}).

\textbf{Pre-training for Intent Recognition.}
Despite the effectiveness of large-scale pre-trained language models \citep{radford2018gpt,devlin2019bert,liu2019roberta,brown2020gpt3}, the inherent mismatch in linguistic behavior between the pre-training datasets and dialogues encourages the research of continual pre-training on dialogue corpus.
Most previous works proposed to pre-train on open domain dialogues in a self-supervised manner \cite{mehri2020dialoglue,wu2020tod,henderson2020convert,ehsan2020simpletod}.
Recently, several works pointed out that pre-training with relavant tasks can be effective for intent recognition.
For example, \citet{zhang2020discriminative} %
formulated intent recognition as a sentence similarity task and 
pre-trained on natural language inference (NLI) datasets.
\citet{vulic2021convfit,zhang2021contrastive} pre-trained with a contrastive loss on intent detection tasks.
Our multi-task pre-training method is inspired from \citet{zhang2021effectiveness} which leverages publicly available intent datasets and unlabeled data in the current domain for pre-training to improve the performance of few-shot intent detection.
However, we argue that the method is more suitable to be applied for NID due to the natural existence of unlabeled utterances.

\textbf{Contrastive Representation Learning.}
Contrastive learning 
has shown promising results in computer vision \citep{bachman2019cpc,chen2020simple,he2019moco,khosla2020supervised} and gained popularity in natural language processing.
Some recent works used unsupervised contrastive learning to learn sentence embeddings  \citep{gao2021simcse,yan2021consert,kim2021guided,giorgi2021declutr}.
Specifically, \citet{gao2021simcse,yan2021consert} showed that contrastive loss can avoid an anisotropic embedding space.
\citet{kim2021guided} proposed a self-guided contrastive training to improve the quality of BERT representations.
\citet{giorgi2021declutr} proposed to pre-train a universal sentence encoder by contrasting a randomly sampled text segment from nearby sentences.
\citet{zhang2021contrastive} demonstrated that self-supervised contrastive pre-training and supervised contrastive fine-tuning can benefit few-shot intent recognition.
\citet{zhang2021supporting} showed that combining a contrastive loss with a clustering objective can improve short text clustering.
Our proposed contrastive loss is tailored for clustering, which encourages utterances with similar semantics to group together and avoids pushing away false negatives as in the conventional contrastive loss.

%% file: sections/method.tex
\makeatletter
\newcommand*{\rom}[1]{\expandafter\@slowromancap\romannumeral #1@}
\makeatother

\begin{figure*}[htp]
    \centering
    \includegraphics[scale=0.50,,trim=4 4 4 2,clip]{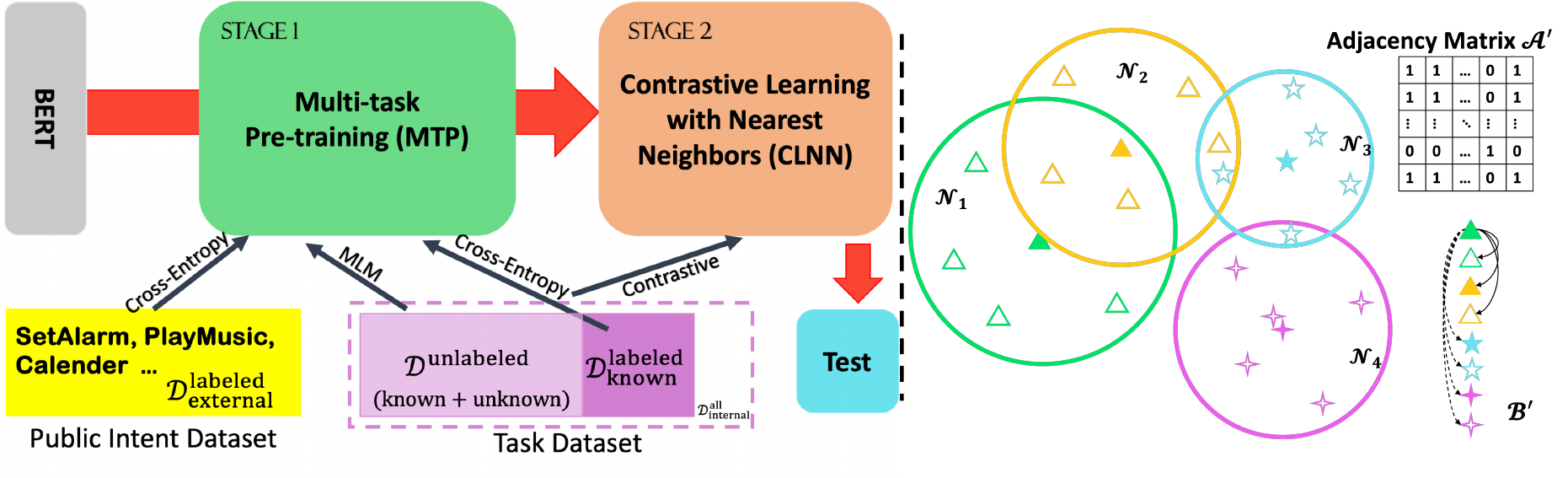}
    \caption{\textbf{The left part shows the overall workflow of our method} where the training order is indicated by the red arrow. The datasets and corresponding loss functions used in each training stage are indicated by the black arrows. \textbf{The right part illustrates a simple example of CLNN}. A batch of four training instances $\{x_i\}_{i=1}^4$ (solid markers) and their respective neighborhoods $\{\mathcal{N}_i\}_{i=1}^4$ are plotted (hollow markers within large circles). Since $x_2$ falls within $\mathcal{N}_1$, $x_2$ along with its neighbors are taken as positive instance for $x_1$ (but not vice versa since $x_1$ is not in $\mathcal{N}_2$).
    We also show an example of adjacency matrix $\bf {A}^{\prime}$ and augmented batch $\mathcal{B}^{\prime}$. The pairwise relationships with the first instance in the batch are plotted with solid lines indicating positive pairs and dashed lines indicating negative pairs.}
    \label{fig:method}
\end{figure*}

\section{Method}

\textbf{Problem Statement.}
To develop an intent recognition model, we usually prepare a set of expected intents $\mathcal{C}_{k}$ along with a few annotated utterances $\mathcal{D}^{\text{labeled}}_{\text{known}}=\{(x_i, y_i)|y_i\in \mathcal{C}_k\}$ for each intent.
After deployed, the system will encounter utterances $\mathcal{D}^{\text{unlabeled}}=\{x_i|y_i\in \{\mathcal{C}_k, \mathcal{C}_u\}\}$ from both pre-defined (known) intents $\mathcal{C}_{k}$ and unknown intents  $\mathcal{C}_{u}$.
The aim of new intent discovery (NID) is to identify the emerging intents $\mathcal{C}_{u}$ in $\mathcal{D}^{\text{unlabeled}}$.
NID can be viewed as a direct extension of out-of-distribution (OOD) detection, where we not only need to identify OOD examples but also discover the underlying clusters.
NID is also different from zero-shot learning in that we do not presume access to any kind of class information during training.
In this work, we consider both unsupervised and semi-supervised NID, which are distinguished by the existence of $\mathcal{D}_{\text{known}}^{\text{labeled}}$, following \citet{zhang2021aligned}.

\textbf{Overview of Our Approach.}
As shown in Fig.~\ref{fig:method}, we propose a two-stage framework that addresses the research questions mentioned in Sec.~\ref{intro}.
In the first stage, we perform multi-task pre-training (MTP) that jointly optimizes a cross-entropy loss on external labeled data and a self-supervised loss on target unlabeled data (Sec.~\ref{method:pki}).
In the second stage, we first mine top-$K$ nearest neighbors of each training instance in the embedding space and then 
perform contrastive learning with nearest neighbors (CLNN) (Sec.~\ref{method:clnn}). 
After training, we employ a simple non-parametric clustering algorithm 
to obtain clustering results. 


\subsection{Stage 1: Multi-task Pre-training (MTP)}
\label{method:pki}
We propose a multi-task pre-training objective that combines a classification task on external data from publicly available intent detection datasets and a self-supervised learning task on internal data from the current domain. Different from previous works~\citep{lin2020constrained,zhang2021aligned}, our pre-training method does not rely on annotated data ($\mathcal{D}^{\text{labeled}}_{\text{known}}$) from the current domain and hence can be applied in an unsupervised setting.
Specifically, we first initialize the model with a pre-trained BERT encoder~\citep{devlin2019bert}. Then, we employ a joint pre-training loss as in \citet{zhang2021effectiveness}. The loss consists of a cross-entropy loss on external labeled data and a masked language modelling (MLM) loss on all available data from the current domain:
\begin{equation}
    \label{eq:stg1}
    \mathcal{L}_{\text{stg1}} = \underbrace{\mathcal{L}_{\text{ce}}(\mathcal{D}^{\text{labeled}}_{\text{external}};\theta)}_\text{supervised}+\underbrace{\mathcal{L}_{\text{mlm}}(\mathcal{D}^{\text{all}}_{{\text{internal}}};\theta)}_{\text{self-supervised}},
\end{equation}
where $\theta$ are model parameters. For the supervised classification task, we leverage an external public intent dataset with diverse domains (e.g., CLINC150~\citep{larson2019clinc}), denoted as $\mathcal{D}^{\text{labeled}}_{\text{external}}$, following~\citet{zhang2021effectiveness}. For the self-supervised MLM task, we use all available data (labeled or unlabeled) from the current domain, denoted as $\mathcal{D}^{\text{all}}_{\text{internal}}$. 

Intuitively, the classification task aims to learn general knowledge of intent recognition with annotated utterances in external intent datasets, while the self-supervised task learns domain-specific semantics with utterances collected in the current domain. Together, they enable learning semantic utterance representations to provide proper cues for the subsequent clustering task. As will be shown in Sec.~\ref{exp:pki}, both tasks are essential for NID.


For semi-supervised NID, we can further utilize the annotated data in the current domain to conduct continual pre-training, by replacing $\mathcal{D}^{\text{labeled}}_{\text{external}}$ in Eq.~\ref{eq:stg1} to $\mathcal{D}^{\text{labeled}}_{\text{known}}$. This step is not included in unsupervised NID.

\subsection{Stage 2: Contrastive Learning with Nearest Neighbors (CLNN)}
\label{method:clnn}
In the second stage, we propose a contrastive learning objective 
that pulls together neighboring instances and pushes away distant ones in the embedding space to learn compact representations for clustering.
Concretely, we first encode the utterances  with the pre-trained model from stage 1. Then, for each utterance $x_i$, we search for its top-$K$ nearest neighbors in the embedding space using inner product as distance metric to form a neighborhood $\mathcal{N}_i$. The utterances in $\mathcal{N}_i$ are supposed to share a similar intent as $x_i$. 
During training, we sample a minibatch of utterances $\mathcal{B}=\{x_i\}_{i=1}^M$.
For each utterance $x_i \in \mathcal{B}$, we uniformly sample one neighbor $x^{\prime}_i$ from its neighborhood $\mathcal{N}_i$. We then use data augmentation to generate $\tilde x_i$ and $\tilde x^{\prime}_i$ for $x_i$ and $x^{\prime}_i$ respectively.
Here, we treat $\tilde x_i$ and $\tilde x^{\prime}_i$ as two views of $x_i$, which form a positive pair.
We then obtain an augmented batch $\mathcal{B}^{\prime}=\{\tilde{x}_i,\tilde{x}^\prime_i\}_{i=1}^{M}$ with all the generated samples.
To compute contrastive loss, we construct an adjacency matrix $\bm {A}^{\prime}$ for $\mathcal{B}^{\prime}$, which is a $2M\times2M$ binary matrix where $1$ indicates positive relation (either being neighbors or having the same intent label in semi-supervised NID) and 0 indicates negative relation.
Hence, we can write the contrastive loss as:
\begin{equation}
\label{eq:contrastive}
    l_{i} = -\frac{1}{|\mathcal{C}_i|}\sum_{j \in \mathcal{C}_i} \log \frac{\exp(\text{sim}(\tilde{h}_i, \tilde{h}_j)/\tau)}{\sum_{{k\neq i}}^{2M} \exp(\text{sim}(\tilde{h}_i, \tilde{h}_k)/\tau)},
\end{equation}
\begin{equation}
    \mathcal{L}_{\text{stg2}} = \frac{1}{2M} \sum_{i=1}^{2M} l_i,
\end{equation}
where $\mathcal{C}_i\equiv\{\bm {A}^{\prime}_{i,j}=1|j\in\{1,...,2M\}\}$ denotes the set of instances having positive relation with $\tilde x_i$ and $|\mathcal{C}_i|$ is the cardinality.
$\tilde{h}_i$ is the embedding for utterance $\tilde{x}_i$.
$\tau$ is the temperature parameter.
$\text{sim}(\cdot,\cdot)$ is a similarity function (e.g., dot product) on a pair of normalized feature vectors. 
During training, the neighborhood will be updated every few epochs. We implement the contrastive loss following~\citet{khosla2020supervised}. 

Notice that the main difference between Eq.~\ref{eq:contrastive} and conventional contrastive loss is how we construct the set of positive instances $\mathcal{C}_i$.
Conventional contrastive loss can be regarded as a special case of Eq.~\ref{eq:contrastive} with neighborhood size $K=0$ and the same instance is augmented twice to form a positive pair \citep{chen2020simple}. After contrastive learning, a 
non-parametric clustering algorithm such as $k$-means can be applied to obtain cluster assignments.



\textbf{Data Augmentation.}
Strong data augmentation has been shown to be beneficial in contrastive learning \citep{chen2020simple}.
We find that it is inefficient to directly apply existing data augmentation methods such as EDA \citep{wei2019eda}, which are designed for general sentence embedding. We observe that the intent of an utterance can be expressed by only a small subset of words such as ``suggest restaurant'' or ``book a flight''.
While it is hard to identify the keywords for an unlabeled utterance, randomly replacing a small amount of tokens in it with some random tokens from the library will not affect intent semantics much. This approach works well in our experiments (See Table~\ref{tab:ablation_aug} RTR).

\textbf{Advantages of CLNN.}
By introducing the notion of neighborhood relationship in contrastive learning, CLNN can 1) pull together similar instances and push away dissimilar ones to obtain  more compact clusters; 2) utilize proximity in the embedding space rather than assigning noisy pseudo-labels~\citep{vangansbeke2020scan}; 3) directly optimize in the feature space rather than clustering logits as in \citet{vangansbeke2020scan}, which has been proven to be more effective by \citet{rebuffi2020lsdc}; and 4) naturally incorporate known intents with the adjacency matrix.

%% file: sections/experiment.tex
\section{Experiment}
\label{experiment}

\subsection{Experimental Details}
\label{exp:detail}

\textbf{Datasets.}
We evaluate our proposed method on three popular intent recognition benchmarks. \textbf{BANKING} (\citealp{casanueva2020efficient}) is a fine-grained dataset with $77$ intents collected from banking dialogues, \textbf{StackOverflow} (\citealp{xu2015short}) is a large scale dataset collected from online queries, \textbf{M-CID} (\citealp{arora2020cross}) is a smaller dataset collected for Covid-19 services. We choose \textbf{CLINC150} (\citealp{larson2019clinc}) as our external public intent dataset in stage 1 
due to its high-quality annotations and coverage of diverse domains. The dataset statistics are summarized in Table~\ref{tab:Dataset statistics}. We use the same splits of BANKING and StackOverflow as in \citet{zhang2021textoir}. Details about dataset splitting are provided in the Appendix.

\begin{table}[t]
\centering
\small
\begin{tabular}{lccc}
\toprule
Dataset & domain  & \#Intents & \#Utterances \\
\midrule
CLINC150       & general & 120   & 18,000 \\
\midrule
BANKING & banking & 77   & 13,083 \\
StackOverflow & questions & 20   & 20,000    \\  
M-CID   & covid-19 & 16   & 1,745    \\  
\bottomrule
\end{tabular}
\caption{Dataset statistics.}
\label{tab:Dataset statistics}
\end{table}

\input{tables/main_un}
\input{tables/main_semi}

\input{tables/fig_scatter}

\input{tables/fig_ablation_mtp}
\input{tables/fig_ablation_clnn}

\textbf{Experimental Setup.}
We evaluate our proposed method on both unsupervised and semi-supervised NID.
\emph{Notice that in unsupervised NID, no labeled utterances from the current domain are provided.}
For clarity, we define two variables. The proportion of known intents is defined as $|\mathcal{C}_k|/(|\mathcal{C}_k|+|\mathcal{C}_u|)$ and referred to as ``\textbf{known class ratio ($\text{KCR}$)}'', and the proportion of labeled examples for each known intent is denoted as ``\textbf{labeled ratio ($\text{LAR}$)}''.
The labeled data are randomly sampled from the original training set.
Notice that, $\text{KCR}=0$ means unsupervised NID, and $\text{KCR} > 0$ means semi-supervised NID.
In the following sections, we provide experimental results for both unsupervised NID and semi-supervised NID with $\text{KCR}=\{25\%,50\%,75\%\}$ and $\text{LAR}=\{10\%,50\%\}$.


\textbf{Evaluation Metrics.}
We adopt three popular evaluation metrics for clustering: normalized mutual information (NMI), adjusted rand index (ARI), and accuracy (ACC).

\textbf{Baselines and Model Variants.}
We summarize the baselines compared in our experiments for both unsupervised and semi-supervised NID.
Our implementation is based on \citet{zhang2021textoir}.\footnote{For fair comparison, the baselines are re-run with TEXTOIR: \url{https://github.com/thuiar/TEXTOIR}, and hence some results are different from those reported in \citet{lin2020constrained,zhang2021aligned}.}

\begin{itemize}
    \item \textbf{Unsupervised baselines.}
(1) GloVe-KM  and (2) GloVe-AG  are based on GloVe \citep{pennington2014glove} embeddings and then evaluated with $k$-means \citep{macqueen1967kmeans} or agglomerative clustering \citep{gowda1984ag} respectively.
(3) BERT-KM applies $k$-means on BERT embeddings.
(4) SAE-KM adopts $k$-means on embeddings of stacked auto-encoder.
(5) Deep Embedding Clustering (SAE-DEC) \citep{xie2016dec} and (6) Deep Clustering Network (SAE-DCN) \citep{bo2017dcn} are unsupervised clustering methods based on stacked auto-encoder.
    \item \textbf{Semi-supervised baselines.}
(1) BERT-KCL \citep{hsu2018kcl} and (2) BERT-MCL \citep{hsu2019mcl} employs pairwise similarity task for semi-supervised clustering.
(3) BERT-DTC \citep{han2019dtc} extends DEC into semi-supervised scenario.
(4) CDAC+ \citep{lin2020constrained} employs a pseudo-labeling process.
(5) Deep Aligned Clustering (DAC) \citep{zhang2021aligned} improves Deep Clustering \citep{caron2018deep} by aligning clusters between iterations.
    \item Our model variants include MTP and MTP-CLNN, which correspond to applying $k$-means on utterance representations learned in stage 1 and stage 2 respectively. Further, we continue to train a DAC model on top of MTP to form a  stronger baseline MTP-DAC for semi-supervised NID.
\end{itemize}



\textbf{Implementation.}
We take pre-trained \textit{bert-base-uncased} model from \citet{wolf2019huggingfaces}\footnote{\url{https://github.com/huggingface/transformers}} as our base model and we use the \textit{[CLS]} token as the BERT representation.
For MTP, we first train until convergence on the external dataset, and then when training on $D^{\text{labeled}}_{\text{known}}$, we use a development set to validate early-stopping with a patience of $20$ epochs following \citet{zhang2021aligned}.
For contrastive learning, we project a 768-d BERT embedding to an 128-d vector with a two-layer MLP and set the temperature as $0.07$.
For mining nearest neighbors, we use the inner product method provided by \citet{johnson2017faiss}\footnote{\url{https://github.com/facebookresearch/faiss}}.
We set neighborhood size $K=50$ for BANKING and M-CID, and $K=500$ for StackOverflow, since we empirically find that the optimal $K$ should be roughly half of the average size of the training set for each class (see Section~\ref{exp:clnn}).
The neighborhood is updated every $5$ epochs.
For data augmentation, the random token replacement probability is set to $0.25$.
For model optimization, we use the AdamW provided by \citet{wolf2019huggingfaces}.
In stage 1, the learning rate is set to $5e^{-5}$. In stage 2, the learning rate is set to $1e^{-5}$ for BANKING and M-CID, and $1e^{-6}$ for StackOverflow.
The batch sizes are chosen based on available GPU memory.
All the experiments are conducted on a single RTX-3090 and averaged over $10$ different seeds.
More details are provided in the Appendix.

\subsection{Result Analysis}
\label{exp:result}
\textbf{Unsupervised NID.}
We show the results for unsupervised NID in Table~\ref{tab:main_un}.
First, comparing the performance of BERT-KM with GloVe-KM and SAE-KM, we observe that BERT embedding performs worse on NID even though it achieves better performance on NLP benchmarks such as GLUE, which manifests learning task-specific knowledge is important for NID.
Second, our proposed pre-training method MTP improves upon baselines by a large margin. Take the NMI score of BANKING for example, MTP outperforms the strongest baseline SAE-DCN by $14.38\%$, which demonstrates the effectiveness of exploiting both external public datasets and unlabeled internal utterances. Furthermore, MTP-CLNN improves upon MTP by around $5\%$ in NMI, $10\%$ in ARI, and $10\%$ in ACC across different datasets.

\textbf{Semi-supervised NID.}
The results for semi-supervised NID are shown in Table~\ref{tab:main_semi}.
First, MTP significantly outperforms the strongest baseline DAC in all settings. For instance, on M-CID, MTP achieves $22.57\%$ improvement over DAC in NMI. Moreover, MTP is less sensitive to the proportion of labeled classes. From $\text{KCR}=75\%$ to $\text{KCR}=25\%$ on M-CID, MTP only drops $8.55\%$ in NMI, as opposed to about $21.58\%$ for DAC. The less performance decrease indicates that our pre-training method is much more label-efficient.
Furthermore, with our proposed contrastive learning, MTP-CLNN consistently outperforms MTP and the combined baseline MTP-DAC. Take BANKING with $\text{KCR}=25\%$ for example, MTP-CLNN improves upon MTP by $4.11\%$ in NMI while surpassing MTP-DAC by $2.63\%$.
A similar trend can be observed when $\text{LAR}=50\%$, and we provide the results in the Appendix.

\textbf{Visualization.}
In Fig.~\ref{fig:scatter}, we show the t-SNE visualization of clusters with embeddings learned by two strongest baselines and our methods. It clearly shows the advantage of our methods, which can produce more compact clusters.
Results on other datasets can be found in the Appendix.

\subsection{Ablation Study of MTP}
\label{exp:pki}
To further illustrate the effectiveness of MTP, we conduct two ablation studies in this section. First, we compare MTP with the pre-training method employed in \citet{zhang2021aligned}, where only internal labeled data are utilized for supervised pre-training (denoted as SUP).\footnote{Notice that we make a simple modification to their pre-training to optimize the entire model rather than the last few layers for fair comparison.}
In Fig.~\ref{fig:ablation}, we show the results of both pre-training methods combined with CLNN with different proportions of known classes. Notice that when $\text{KCR}=0$ there is no pre-training at all for SUP-CLNN. It can be seen that
MTP-CLNN consistently outperforms SUP-CLNN.
Furthermore, the performance gap increases while $\text{KCR}$ decreases, and the largest gap is achieved when $\text{KCR}=0$.
This shows the high effectiveness of our method in data-scarce scenarios.

\input{tables/ablation_mtp}
Second, we decompose MTP into two parts: supervised pre-training on external public data (PUB) and self-supervised pre-training on internal unlabeled data (MLM). We report the results of the two pre-training methods combined with CLNN as well as MTP in Table~\ref{tab:ablation_mtp}. We can easily conclude that either PUB or MLM is indispensable and multi-task pre-training is beneficial.

\subsection{Analysis of CLNN}
\label{exp:clnn}

\textbf{Number of Nearest Neighbors.}
We conduct an ablation study on neighborhood size $K$ in Fig.~\ref{fig:clnn_analysis}.
We can make two main observations.
First, although the performance of MTP-CLNN varies with different $K$, it still significantly outperforms MTP (dashed horizontal line) for a wide range of $K$.
For example, MTP-CLNN is still better than MTP when $K=50$ on StackOverflow or $K=200$ on BANKING. Second, despite the difficulty to search for $K$ with only unlabeled data, we empirically find an effective estimation method, i.e. \textit{to choose $K$ as half of the average size of the training set for each class}\footnote{We presume prior knowledge of the number of clusters. There are some off-the-shelf methods that can be directly applied in the embedding space to determine the optimal number of clusters \citep{zhang2021aligned}.}.
It can be seen that the estimated $K\approx60$ on BANKING and $K\approx40$ on M-CID (vertical dashed lines) lie in the optimal regions, which shows the effectiveness of our empirical estimation method.

\textbf{Exploration of Data Augmentation.}
We compare Random Token Replacement (RTR) used in our experiments with other methods. For instance, dropout is applied on embeddings to provide data augmentation in \citet{gao2021simcse}, randomly shuffling the order of input tokens is proven to be effective in \citet{yan2021consert}, and EDA \citep{wei2019eda} is often applied in text classification.
Furthermore, we compare with a Stop-words Replacement (SWR) variant that only replaces the stop-words with other random stop-words so it minimally affects the intents of utterances.
The results in Table~\ref{tab:ablation_aug} demonstrate that (1) RTR and SWR consistently outperform others, which verifies our hypothesis in Section~\ref{method:clnn}. (2) Surprisingly, RTR and SWR perform on par with each other.
For simplicity, we only report the results with RTR in the main experiments.
\input{tables/ablation_aug}

%% file: tables/main_un.tex
\begin{table*}[ht!]
\centering
\scalebox{0.85}{
\begin{tabular}{cc|ccc|ccc|ccc}
\specialrule{0.1em}{0.2em}{0.2em}
\multicolumn{2}{c}{}&\multicolumn{3}{c}{BANKING}&\multicolumn{3}{c}{StackOverflow}&\multicolumn{3}{c}{M-CID}\\
&Methods & NMI & ARI & ACC & NMI & ARI & ACC & NMI & ARI & ACC\\
\specialrule{0.05em}{0.1em}{0.2em}
\multirow{8}{*}{unsupervised} 
                    &GloVe-KM& 48.75&12.74&27.92 & 21.79&4.54&24.26 & 46.40&35.57&46.99   \\
                    &GloVe-AG& 52.76&14.41&31.18 & 23.45&4.85&24.48 & 51.23&32.57&42.35  \\
                    &SAE-KM& 60.12&24.00&37.38 & 48.72&23.36&37.16 & 51.03&43.51&52.95 \\
                    &SAE-DEC& 62.92&25.68&39.35 & 61.32&21.17&57.09 & 50.69&44.52&53.07 \\
                    &SAE-DCN& 62.94&25.69&39.36 & 61.34&34.98&57.09 & 50.69&44.52&53.07 \\
                    &BERT-KM& 36.38&5.38&16.27 & 11.60&1.60&13.85 & 37.37&14.02&33.81 \\
                    & MTP \textbf{(Ours)} & 77.32&47.33&57.99 & 63.85&48.71&66.18 & 72.40&53.04&68.94 \\
                    & MTP-CLNN \textbf{(Ours)} & \bf81.80&\bf55.75&\bf65.90 & \bf78.71&\bf67.63&\bf81.43 & \bf79.95&\bf66.71&\bf79.14 \\
\specialrule{0.1em}{0.1em}{0.2em}
\end{tabular}
}
\caption{\label{tab:main_un}
Performance on unsupervised NID. For each dataset, the best results are marked in bold.
}
\end{table*}

%% file: tables/main_semi.tex
\begin{table*}[ht!]
\centering
\scalebox{0.85}{
\begin{tabular}{cc|ccc|ccc|ccc}
\specialrule{0.1em}{0.2em}{0.2em}
\multicolumn{2}{c}{}&\multicolumn{3}{c}{BANKING}&\multicolumn{3}{c}{StackOverflow}&\multicolumn{3}{c}{M-CID}\\
KCR&Methods & NMI & ARI & ACC & NMI & ARI & ACC & NMI & ARI & ACC\\
\specialrule{0.05em}{0.1em}{0.2em}
\multirow{8}{*}{25\%} 
                    &BERT-DTC&56.05 & 20.19 &32.91 &33.38  &16.45 &30.32 &36.00 &13.64 &27.51   \\
                    &BERT-KCL&53.85 & 20.07 &28.79 &35.47  &16.80 &32.88 &29.35 &11.58 &24.76  \\
                    &BERT-MCL&49.46 & 15.51 &24.53 &29.44 &14.99 &31.50 &31.16 &11.30 & 26.13 \\
                    &CDAC+&67.65& 34.88& 48.79 &74.33 &39.44 &74.30 &43.89& 19.65& 39.37\\
                    &DAC&69.85& 37.16 & 49.67 &53.97 &36.46 &53.96 & 49.83& 27.21& 43.72  \\
                    & MTP \textbf{(Ours)} & 80.00&51.86&62.75 & 73.75&61.06&75.98 & 72.40&53.04&68.94 \\
                    & MTP-DAC \textbf{(Comb)} & 81.48&55.64&66.12 & 77.22&61.42&78.60 & 77.79&62.88&77.02 \\
                    & MTP-CLNN \textbf{(Ours)} & \bf84.11&\bf61.29&\bf71.43 & \bf79.68&\bf70.17&\bf83.77 & \bf80.24&\bf66.77&\bf79.20 \\
\specialrule{0.05em}{0.1em}{0.2em}
\multirow{8}{*}{50\%} 
                    &BERT-DTC& 69.68&35.98&48.87 & 53.94&36.79&51.78 & 51.90&28.94&44.70\\
                    &BERT-KCL& 62.86&30.16&40.81 & 57.63&41.90&56.58 & 42.48&22.83&38.11\\
                    &BERT-MCL& 62.50&29.80&42.28 & 49.49&35.96&53.16 & 41.50&21.46&37.99\\
                    &CDAC+& 70.62&38.61&51.97 & 76.18&41.92&76.30 & 50.47&26.01&46.65\\
                    &DAC& 76.41&47.28&59.32 & 70.78&56.44&73.76 & 63.27&43.52&57.19 \\
                    &MTP \textbf{(Ours)}& 82.92&58.46&68.29 & 77.11&66.45&79.28 & 72.40&53.04&68.94 \\
                    &MTP-DAC \textbf{(Comb)}& 83.43&59.78&70.42 & 78.91&67.37&81.27 & 78.17&63.41&77.68 \\
                    &MTP-CLNN \textbf{(Ours)}& \bf85.62&\bf64.93&\bf75.23 & \bf81.03&\bf73.02&\bf85.64 & \bf79.48&\bf65.71&\bf77.85 \\
\specialrule{0.05em}{0.1em}{0.2em}
\multirow{8}{*}{75\%} 
                    &BERT-DTC& 74.51&44.57&57.34 & 67.02&55.14&71.14 & 60.82&38.62&55.42 \\
                    &BERT-KCL& 72.18&44.29&58.70 & 70.38&57.98&71.50 & 54.22&34.60&52.15 \\
                    &BERT-MCL& 74.41&48.08&61.57 & 67.72&55.78&70.82 & 51.33&31.22&50.77 \\
                    &CDAC+& 71.76&40.68&53.46 & 76.68&43.97&75.34 & 55.06&32.52&53.70 \\
                    &DAC& 79.99&54.57&65.87 & 75.31&60.02&78.84 & 71.41&54.22&69.11 \\
                    &MTP \textbf{(Ours)}& 85.17&64.37&74.20 & 80.70&71.68&83.74 & 80.95&69.27&80.92 \\
                    &MTP-DAC \textbf{(Comb)}& 85.78&65.28&75.43 & 80.89&71.17&84.20 & 80.94&68.27&80.89 \\
                    &MTP-CLNN \textbf{(Ours)}& \bf87.52&\bf70.00&\bf79.74 & \bf82.56&\bf75.66&\bf87.63 & \bf83.75&\bf73.22&\bf84.36 \\
\specialrule{0.1em}{0.1em}{0.2em}
\end{tabular}
}
\caption{\label{tab:main_semi}
Performance on semi-supervised NID with different known class ratio. The $\text{LAR}$ is set to $10\%$. For each dataset, the best results are marked in bold. \textbf{Comb} denotes the  baseline method combined with our proposed MTP.
}
\end{table*}

%% file: tables/fig_scatter.tex
\begin{figure*}[t]
    \begin{subfigure}{.24\linewidth}
    \centering
        \includegraphics[scale=0.16]{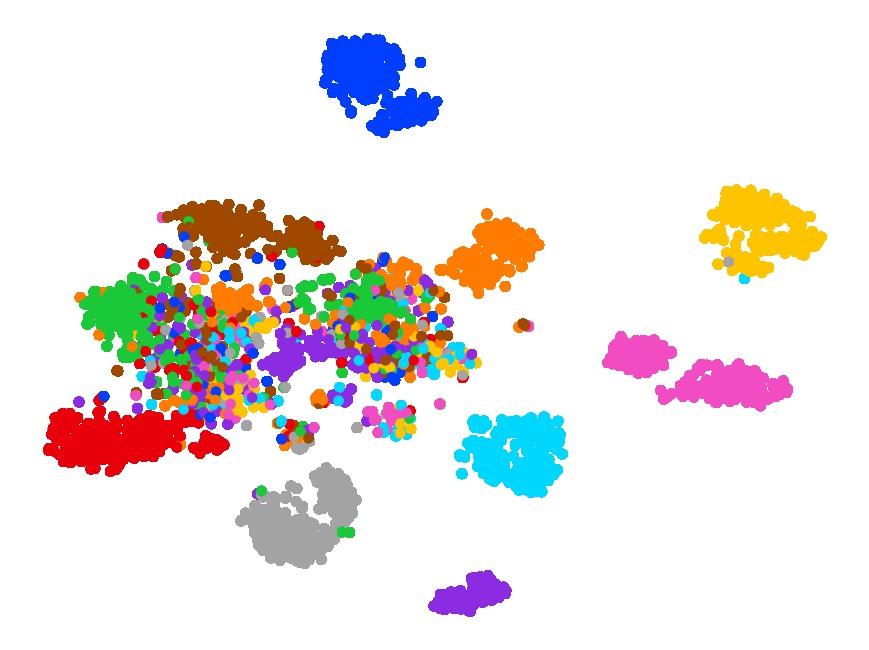}
        \caption{CDAC+}
        \label{subfigure: scatter_bert}
    \end{subfigure}
    \begin{subfigure}{.24\linewidth}
    \centering
        \includegraphics[scale=0.16]{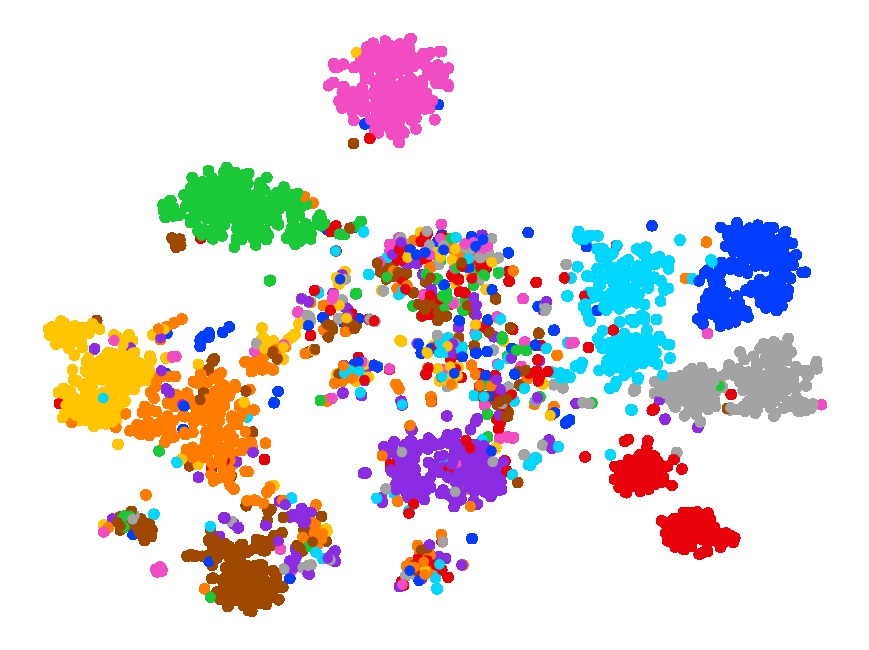}
        \caption{DAC}
        \label{subfigure: scatter_dac}
    \end{subfigure}
    \begin{subfigure}{.24\linewidth}
    \centering
        \includegraphics[scale=0.16]{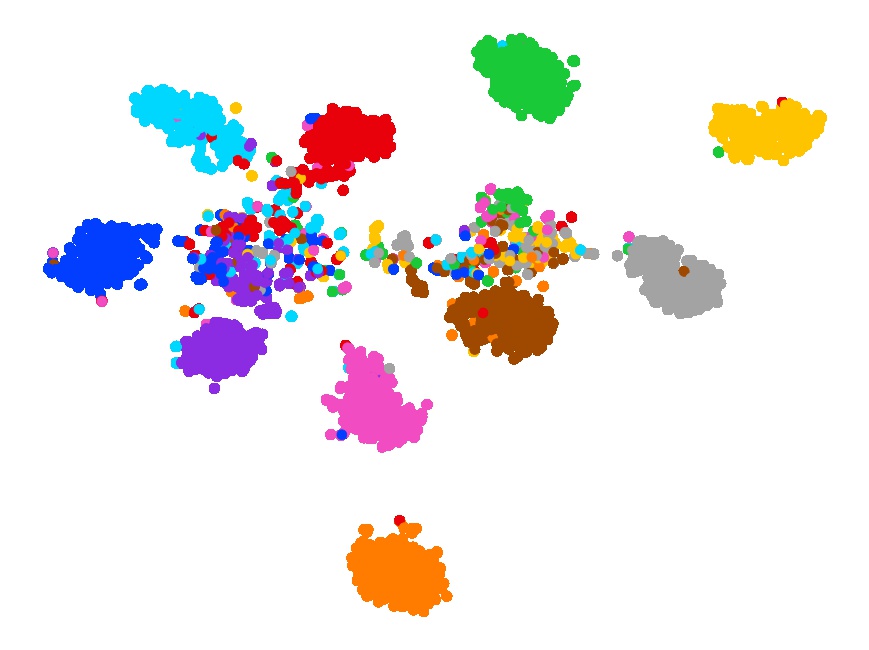}
        \caption{MTP (Ours)}
        \label{subfigure: scatter_pki}
    \end{subfigure}
    \begin{subfigure}{.24\linewidth}
    \centering
        \includegraphics[scale=0.16]{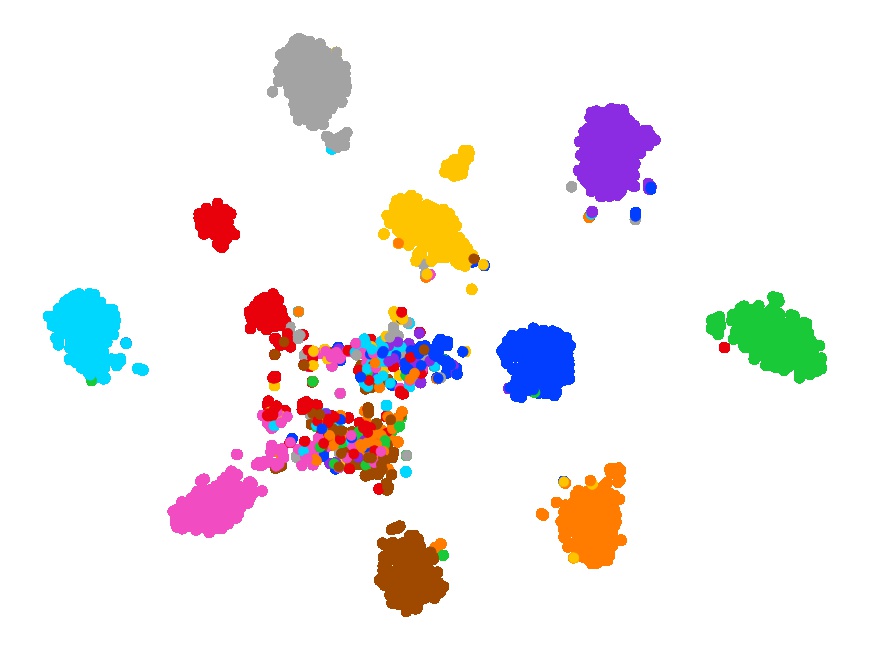}
        \caption{MTP-CLNN (Ours)}
        \label{subfigure: scatter_clnn}
    \end{subfigure}
    \caption{Visulization of embeddings on StackOverflow. $\text{KCR}=25\%$, $\text{LAR}=10\%$. Best viewed in color.}
    \label{fig:scatter}
\end{figure*}

%% file: tables/fig_ablation_mtp.tex
\begin{figure*}[t]
    \begin{subfigure}{.33\linewidth}
    \centering
        \includegraphics[scale=0.20]{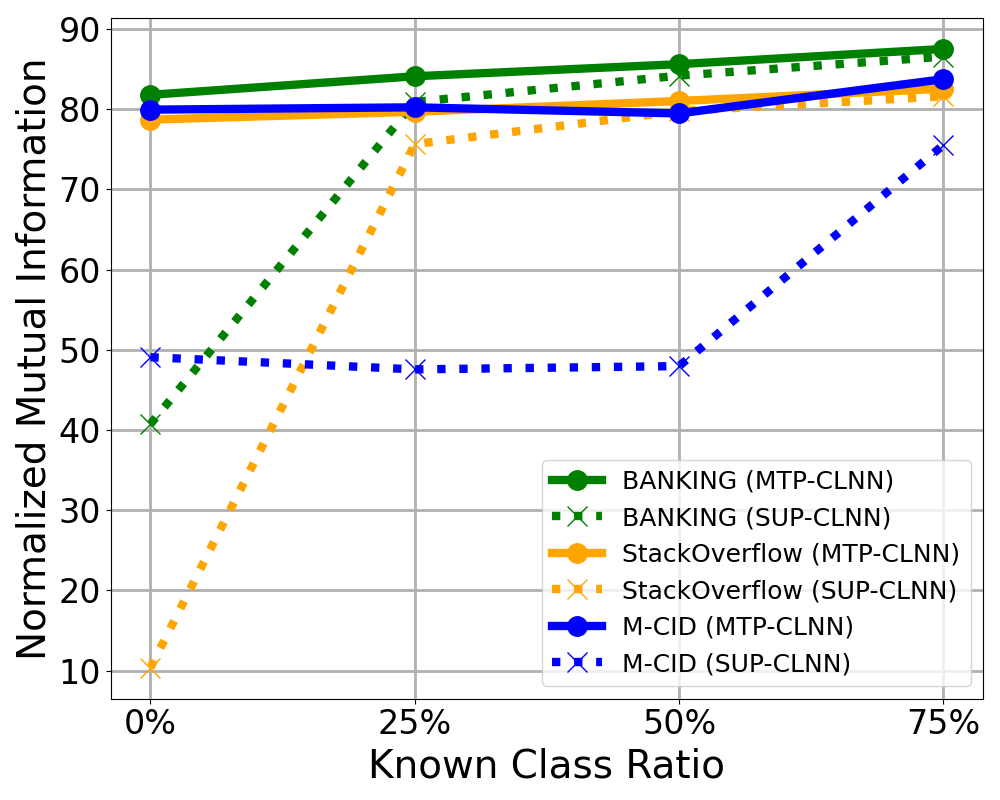}
        \caption{}
        \label{subfigure: ablation_gdp_nmi}
    \end{subfigure}
    \begin{subfigure}{.33\linewidth}
    \centering
        \includegraphics[scale=0.20]{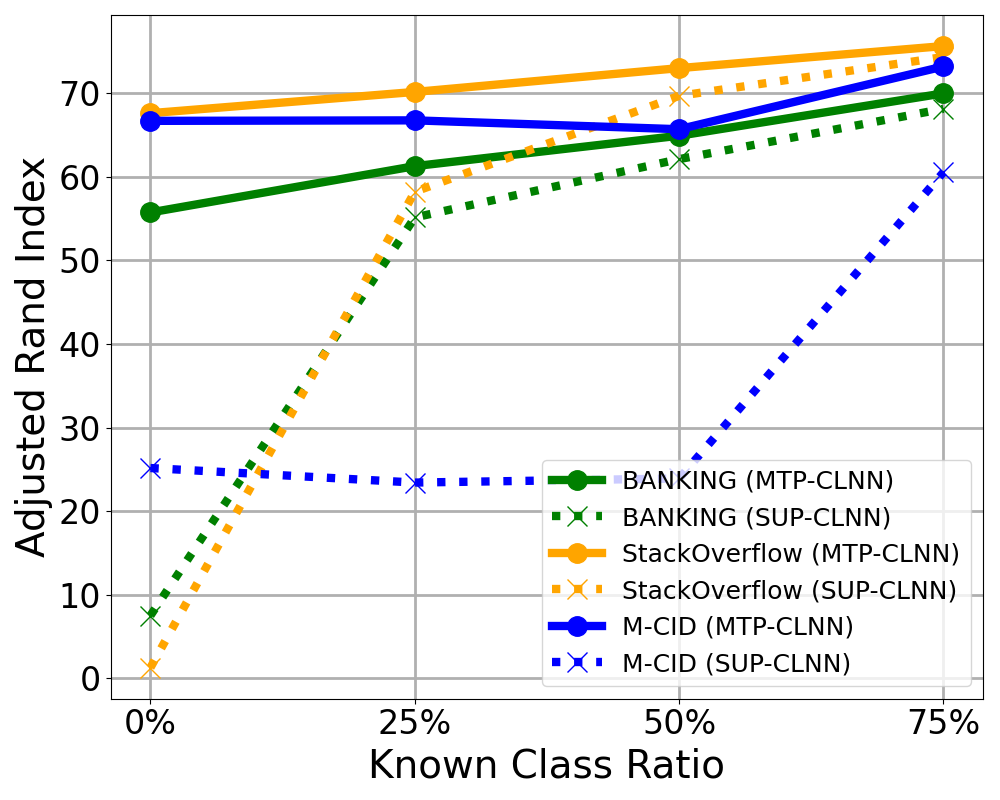}
        \caption{}
        \label{subfigure: ablation_gdp_ari}
    \end{subfigure}
    \begin{subfigure}{.33\linewidth}
    \centering
        \includegraphics[scale=0.20]{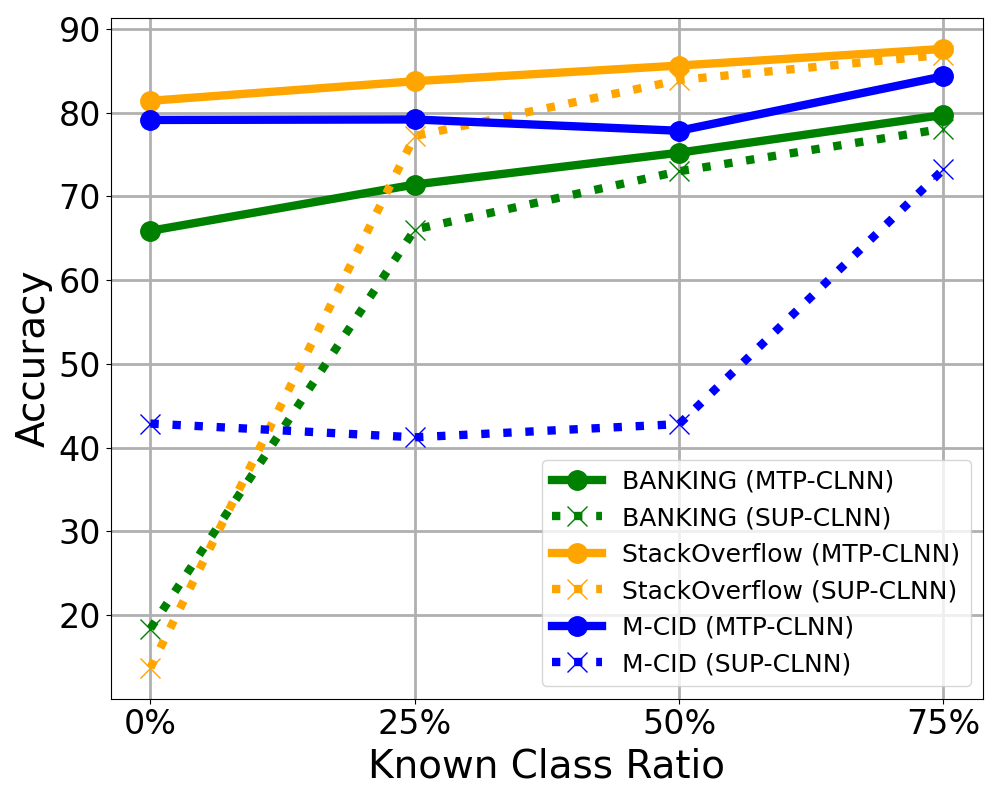}
        \caption{}
        \label{subfigure: ablation_gdp_acc}
    \end{subfigure}
    \caption{Ablation study on the effectiveness of MTP. The $\text{LAR}$ is set to 10\%. SUP stands for supervised pre-training on internal labeled data only. The three columns correspond to results in the three metrics respectively.}
    \label{fig:ablation}
\end{figure*}

%% file: tables/fig_ablation_clnn.tex
\begin{figure*}[t]
    \begin{subfigure}{.33\linewidth}
    \centering
        \includegraphics[scale=0.20]{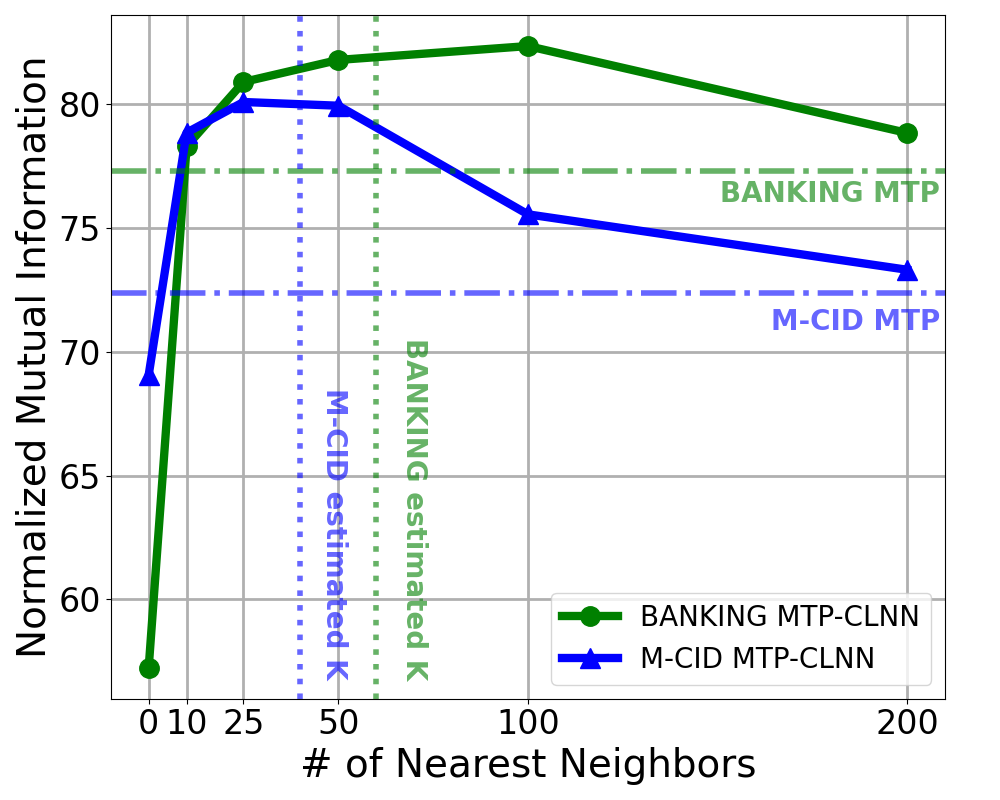}
        \caption{}
        \label{subfigure: ablation_k_nmi}
    \end{subfigure}
    \begin{subfigure}{.33\linewidth}
    \centering
        \includegraphics[scale=0.20]{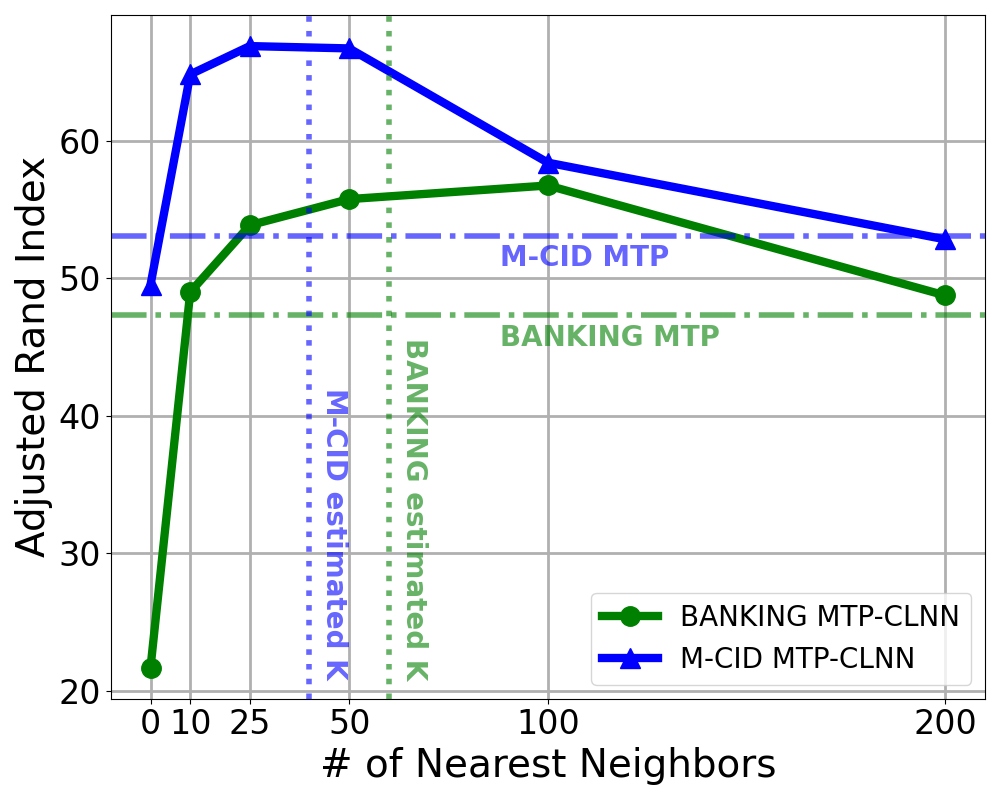}
        \caption{}
        \label{subfigure: ablation_k_ari}
    \end{subfigure}
    \begin{subfigure}{.33\linewidth}
    \centering
        \includegraphics[scale=0.20]{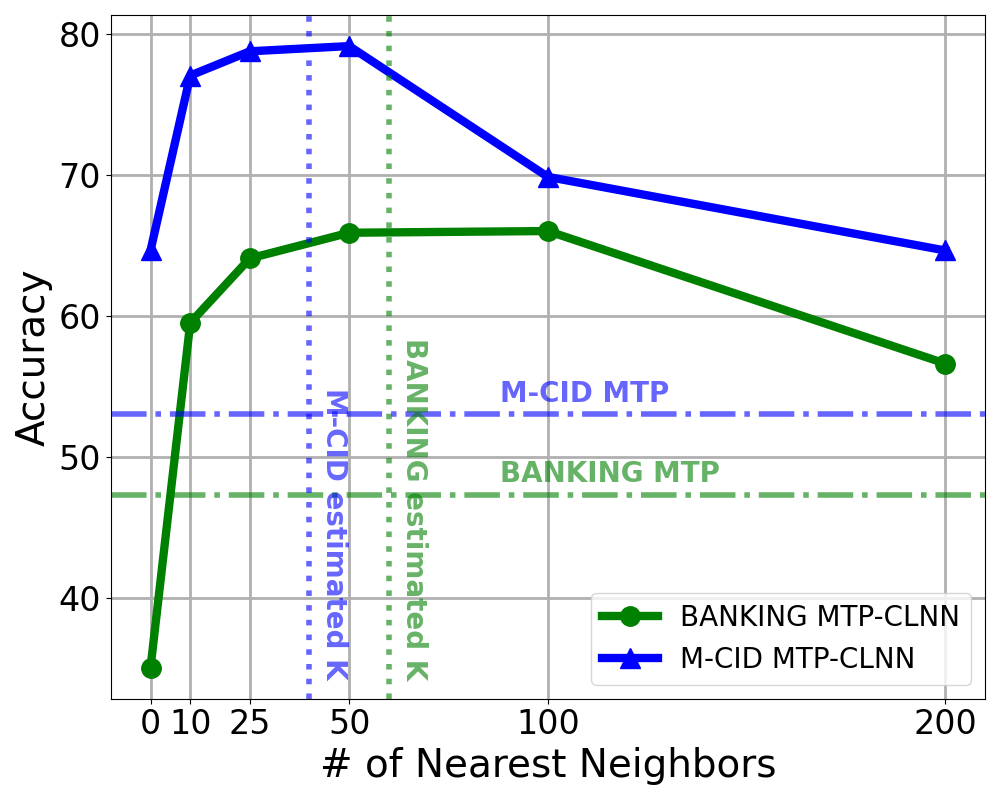}
        \caption{}
        \label{subfigure: ablation_k_acc}
    \end{subfigure}\\
    \begin{subfigure}{.33\linewidth}
    \centering
        \includegraphics[scale=0.20]{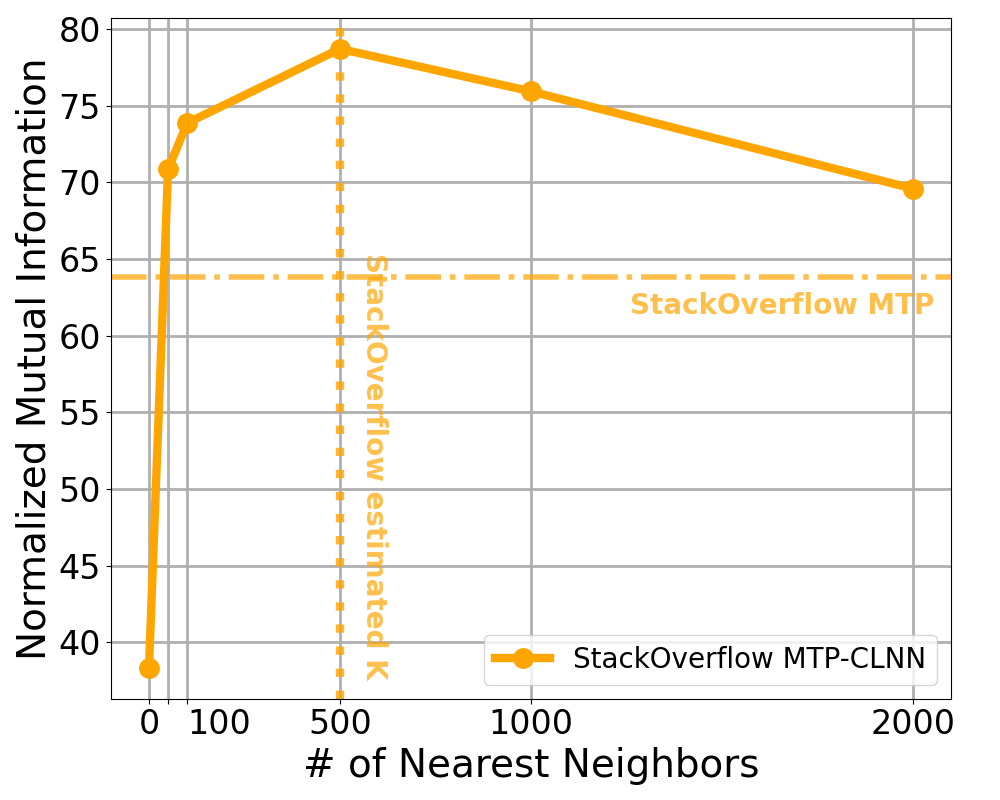}
        \caption{}
        \label{subfigure: ablation_k_so_nmi}
    \end{subfigure}
    \begin{subfigure}{.33\linewidth}
    \centering
        \includegraphics[scale=0.20]{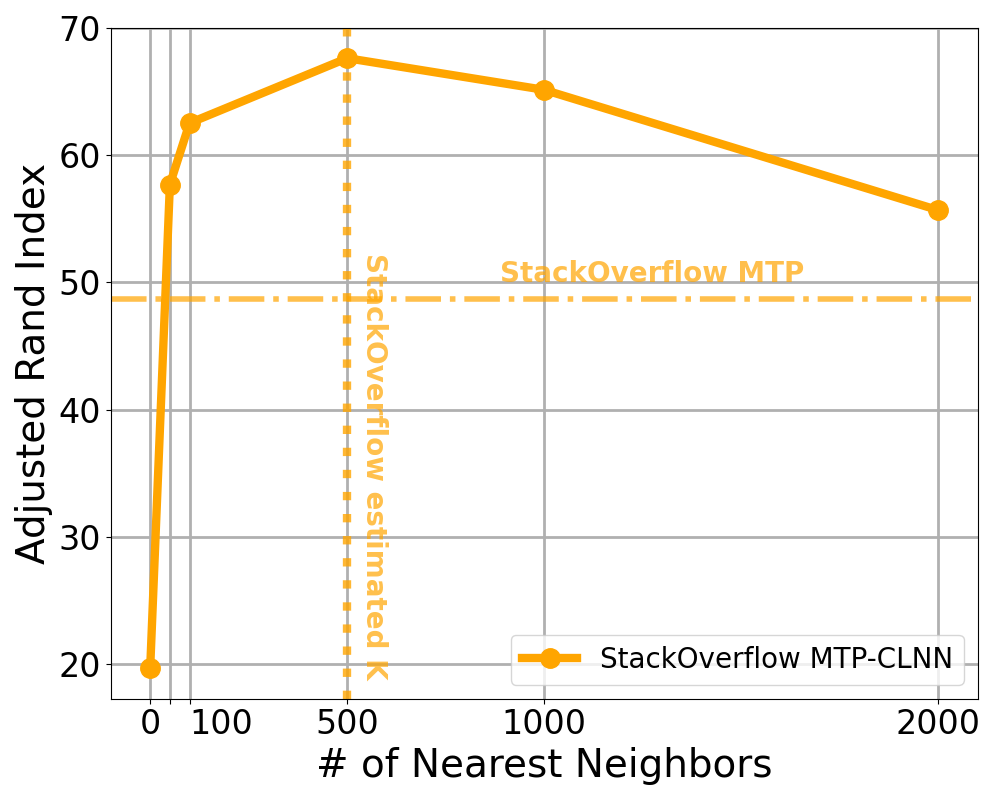}
        \caption{}
        \label{subfigure: ablation_k_so_ari}
    \end{subfigure}
    \begin{subfigure}{.33\linewidth}
    \centering
        \includegraphics[scale=0.20]{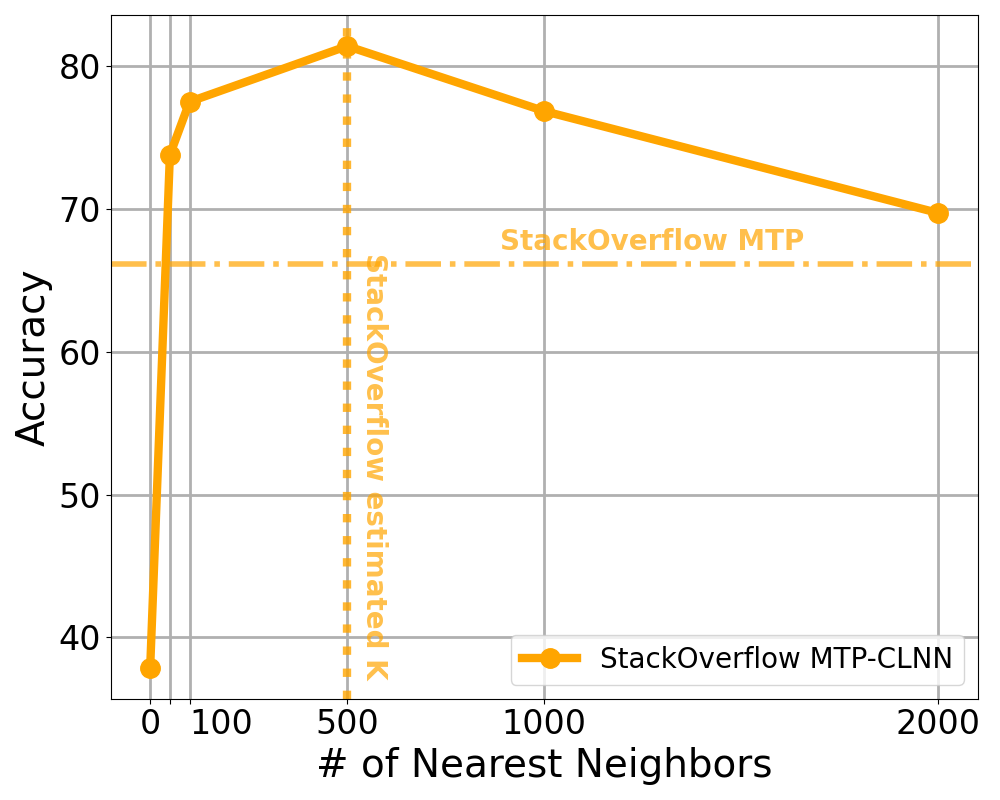}
        \caption{}
        \label{subfigure: ablation_k_so_acc}
    \end{subfigure}
    \caption{Analysis on the number of nearest neighbors in CLNN for unsupervised NID. Vertical dashed lines correspond to our empirical estimations of optima. Horizontal dashed lines represent the results of only training with MTP. When the number of nearest neighbors is $0$, we simply augment the same instance twice as in conventional contrastive learning~\cite{chen2020simple}. The three columns correspond to results in the three metrics respectively.}
    \label{fig:clnn_analysis}
\end{figure*}

%% file: tables/ablation_mtp.tex
\begin{table}[t]
\centering
\scalebox{0.80}{
\setlength\tabcolsep{3.5pt}
\begin{tabular}{c|cc|cc|cc}
\specialrule{0.1em}{0.2em}{0.2em}
\multicolumn{1}{c}{}&\multicolumn{2}{c}{BANKING}&\multicolumn{2}{c}{StackOverflow}&\multicolumn{2}{c}{M-CID}\\
Methods & NMI & ARI & NMI & ARI & NMI & ARI\\
\specialrule{0.05em}{0.1em}{0.2em}
                    PUB-CLNN& 75.69&44.58 & 42.22&24.77 & 73.64&56.94 \\
                    MLM-CLNN& 77.02&47.79 & 78.62&\bf68.77 & 71.28&53.28 \\
                    MTP-CLNN& \bf81.80&\bf55.75 & \bf78.71&67.63 & \bf79.95&\bf66.71\\
\specialrule{0.1em}{0.1em}{0.2em}
\end{tabular}
}
\caption{\label{tab:ablation_mtp}
Ablation study of MTP for unsupervised NID.
}
\end{table}

%% file: tables/ablation_aug.tex
\begin{table}[t]
\centering
\scalebox{0.80}{
\setlength\tabcolsep{3.5pt}
\begin{tabular}{c|cc|cc|cc}
\specialrule{0.1em}{0.2em}{0.2em}
\multicolumn{1}{c}{}&\multicolumn{2}{c}{BANKING}&\multicolumn{2}{c}{StackOverflow}&\multicolumn{2}{c}{M-CID}\\
Methods & NMI & ARI & NMI & ARI & NMI & ARI\\
\specialrule{0.05em}{0.1em}{0.2em}
                    dropout&  79.52&50.83 & 75.60&57.67 & 79.64&66.14 \\
                    shuffle& 79.02&49.72 & 75.70&58.95 & 79.68&66.09 \\
                    EDA& 78.29&49.02 & 71.50&49.80 & 79.73&66.39 \\
                    SWR(Ours) & 82.03&56.18 & 78.48&67.15 & 79.23&65.74 \\
                    RTR(Ours)* & 81.80&55.75 & 78.71&67.63 & 79.95&66.71 \\
\specialrule{0.1em}{0.1em}{0.2em}
\end{tabular}
}
\caption{\label{tab:ablation_aug}
Ablation study on data augmentation for unsupervised NID. * is the method used in the main results.
}
\end{table}

%% file: sections/conclusion.tex
\section{Conclusion}
We have provided simple and effective solutions for two fundamental research questions for new intent discovery (NID): (1) how to learn better utterance representations to provide proper cues for clustering and (2) how to better cluster utterances in the representation space.
In the first stage, we use a multi-task pre-training strategy to exploit both external and internal data for representation learning.
In the second stage, we perform contrastive learning with mined nearest neighbors to exploit self-supervisory signals in the representation space.
Extensive experiments on three intent recognition benchmarks show that our approach can significantly improve the performance of NID in both unsupervised and semi-supervised scenarios.

There are two limitations of this work.
(1) We have only evaluated on balanced data. However, in real-world applications, most datasets are highly imbalanced.
(2) The discovered clusters lack interpretability. Our clustering method can only assign a cluster label to each unlabeled utterance but cannot generate a valid intent name for each cluster.


\section{Acknowledgments}

We would like to thank the anonymous reviewers for their valuable comments. This research was supported by the grants of HK ITF UIM/377 and PolyU DaSAIL project P0030935 funded by RGC.

%% file: appendix.tex
\appendix

\section{Experimental Details}
\label{sec:exp}

\subsection{Datasets}
In this section, we provide more details about the datasets. The development sets are prepared to exclude no unknown intents.
\begin{itemize}
    \item \textbf{BANKING} (\citealp{casanueva2020efficient}) is a fine-grained intent detection dataset in which 77 intents are collected for banking dialogue system. The dataset is splitted into 9,003, 1,000 and 3,080 for training, validation, and test sets respectively.
    \item \textbf{StackOverflow} (\citealp{xu2015short}) is a large scale dataset for online questioning which contains 20 intents with 1,000 examples in each class. We split the dataset into 18,000 for training, 1,000 for validation, and 1,000 for test.
    \item \textbf{M-CID} (\citealp{arora2020cross}) is a small scale dataset for cross-lingual Covid-19 queries. We only use the English subset of this dataset which has 16 intents. We split the dataset into 1,220 for training, 176 for validation, and 349 for test.
    \item \textbf{CLINC150} (\citealp{larson2019clinc}) consists of 10 domains across multiple unique services. We use 8 domains \footnote{The domains ``Banking'' and ``Credit Cards'' are excluded because they are semantically close to the evaluation data.} and remove the out-of-scope data. We only use this dataset during training stage 1.
\end{itemize}

\subsection{Implementation}
The batch size is set to 64 for stage 1 and 128 for stage 2 in all experiments to fully utilize the GPU memory.
In stage 1, we first train until convergence on external data and then train with validation on internal data.
In stage 2, we train until convergence without early-stopping.

\subsection{More Experimental Results}
\begin{figure*}[h]
    \begin{subfigure}{.24\linewidth}
    \centering
        \includegraphics[scale=0.16]{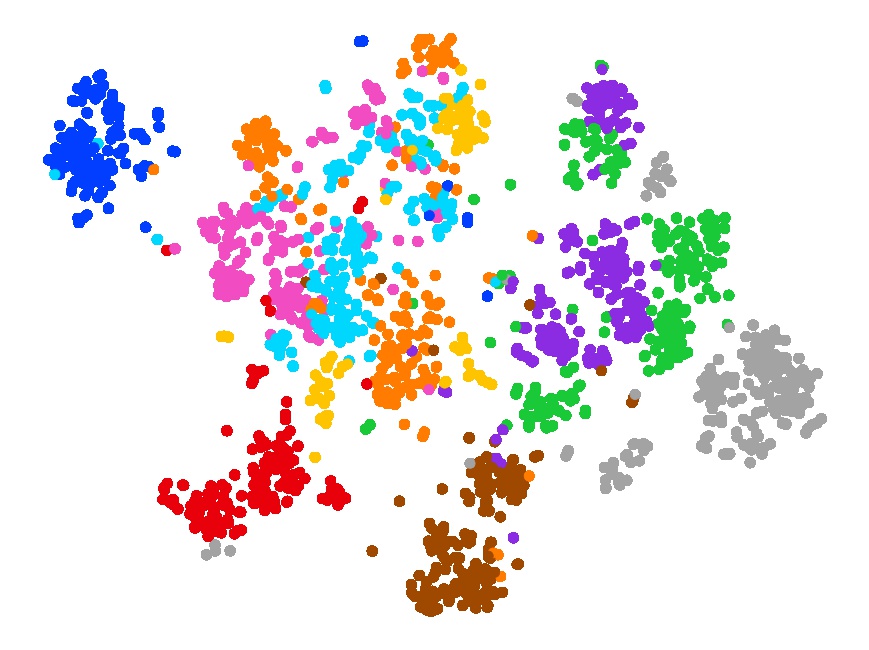}
        \caption{CDAC+}
        \label{subfigure: scatter_bert2}
    \end{subfigure}
    \begin{subfigure}{.24\linewidth}
    \centering
        \includegraphics[scale=0.16]{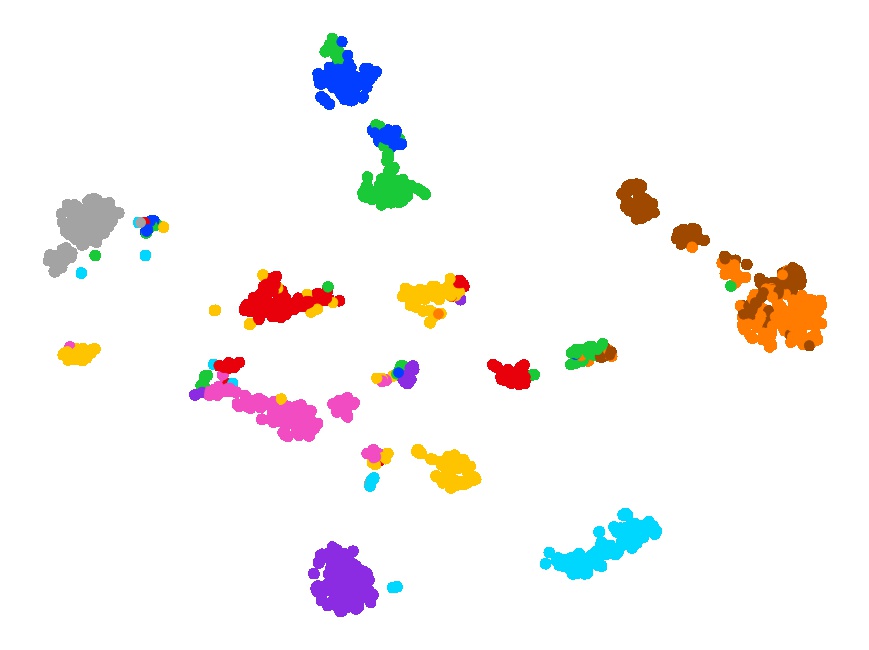}
        \caption{DAC}
        \label{subfigure: scatter_dac2}
    \end{subfigure}
    \begin{subfigure}{.24\linewidth}
    \centering
        \includegraphics[scale=0.16]{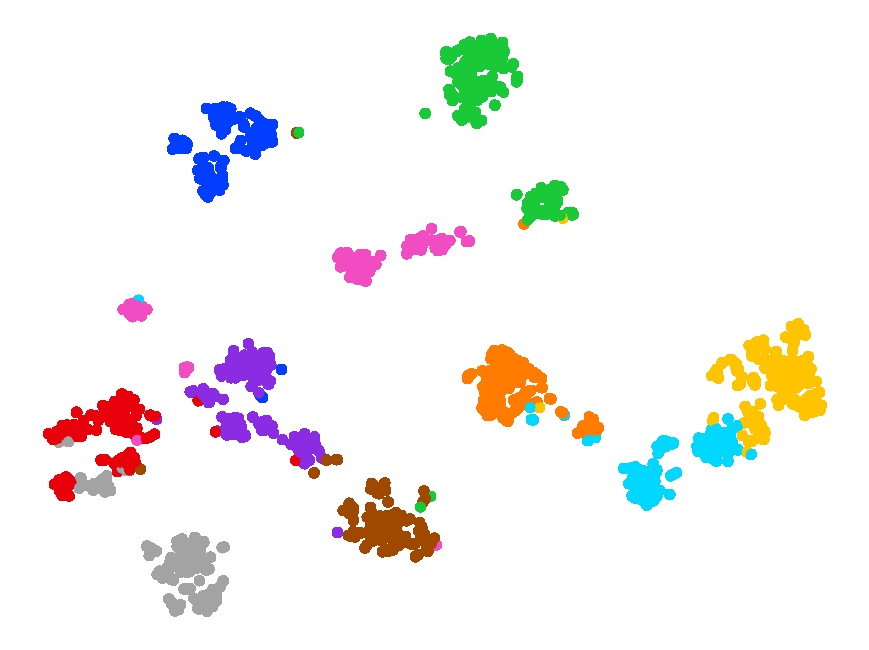}
        \caption{MTP (Ours)}
        \label{subfigure: scatter_pki2}
    \end{subfigure}
    \begin{subfigure}{.24\linewidth}
    \centering
        \includegraphics[scale=0.16]{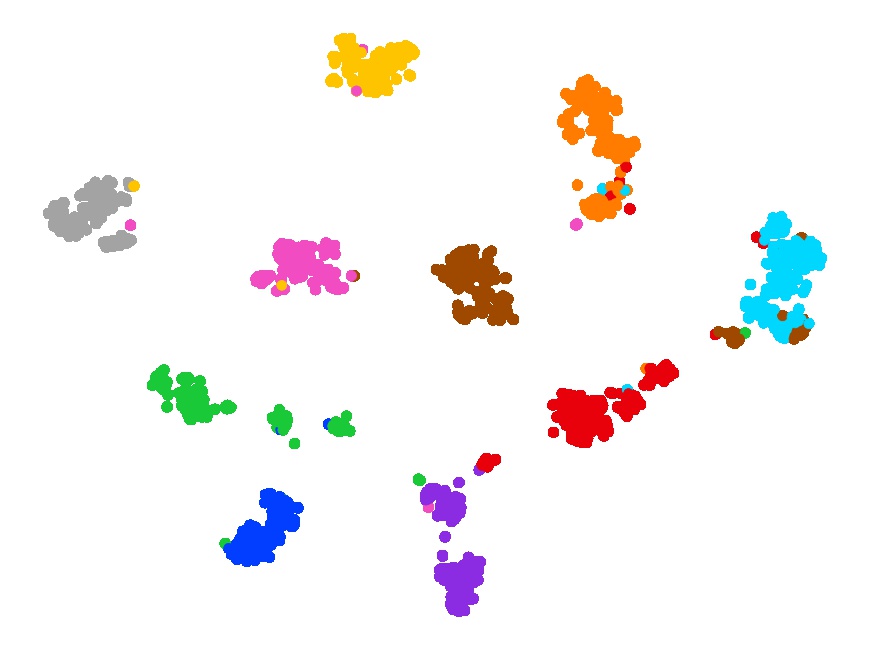}
        \caption{MTP-CLNN (Ours)}
        \label{subfigure: scatter_clnn2}
    \end{subfigure}
    \caption{t-SNE visulization of embeddings on BANKING. $\text{KCR}=25\%$, $\text{LAR}=10\%$. Best viewed in color.}
    \label{fig:scatter2}
\end{figure*}

\begin{figure*}[h]
    \begin{subfigure}{.24\linewidth}
    \centering
        \includegraphics[scale=0.16]{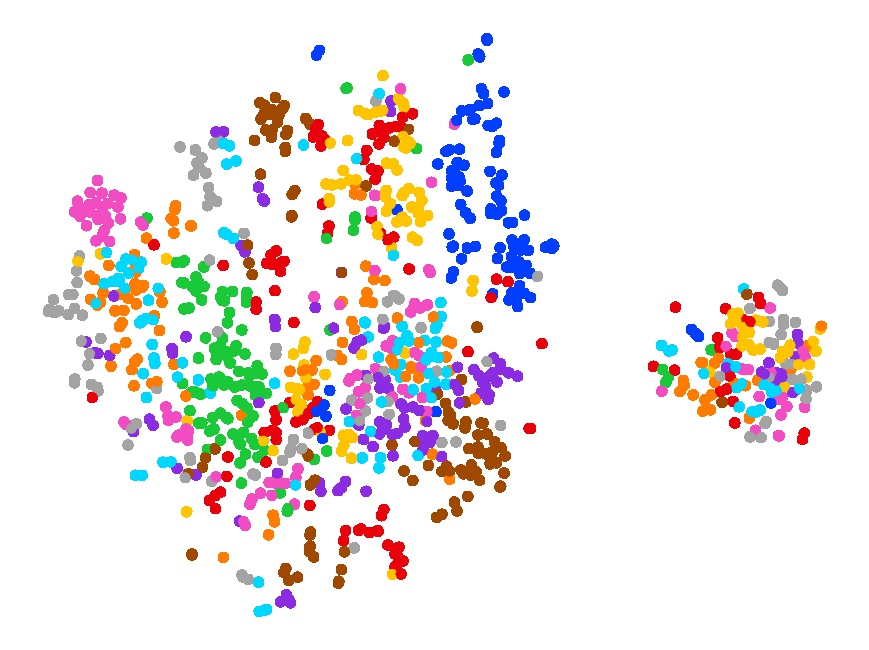}
        \caption{CDAC+}
        \label{subfigure: scatter_bert3}
    \end{subfigure}
    \begin{subfigure}{.24\linewidth}
    \centering
        \includegraphics[scale=0.16]{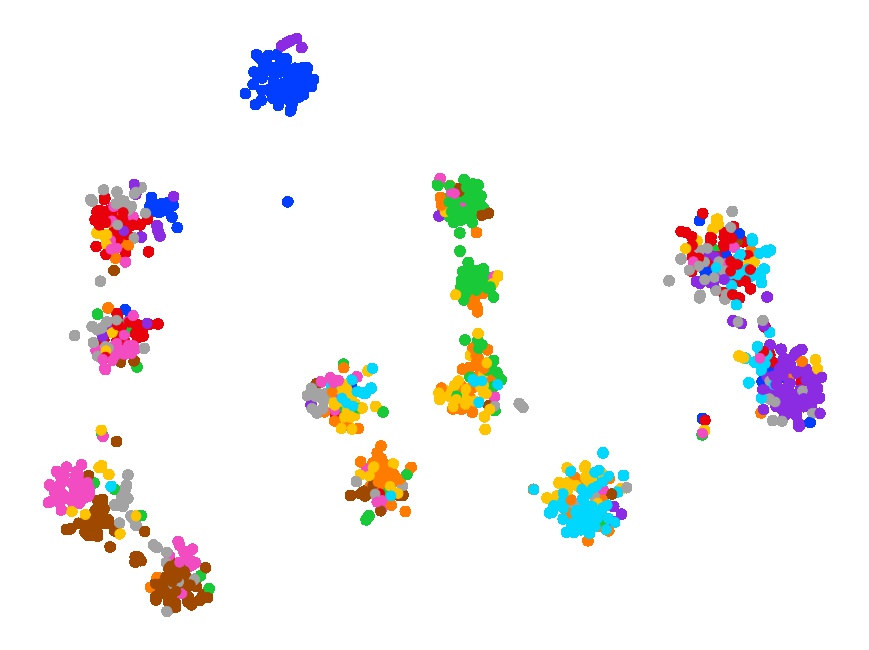}
        \caption{DAC}
        \label{subfigure: scatter_dac3}
    \end{subfigure}
    \begin{subfigure}{.24\linewidth}
    \centering
        \includegraphics[scale=0.16]{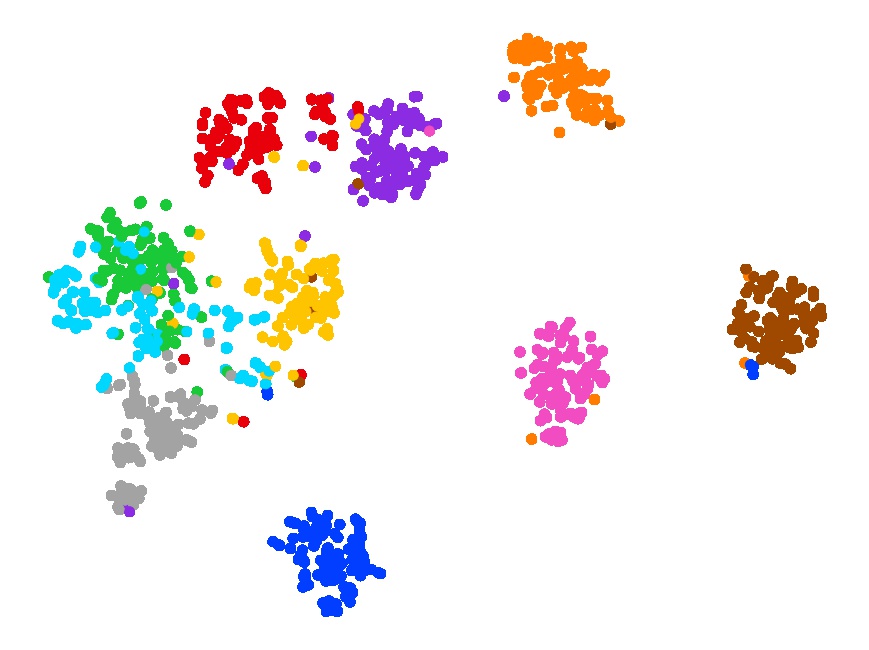}
        \caption{MTP (Ours)}
        \label{subfigure: scatter_pki3}
    \end{subfigure}
    \begin{subfigure}{.24\linewidth}
    \centering
        \includegraphics[scale=0.16]{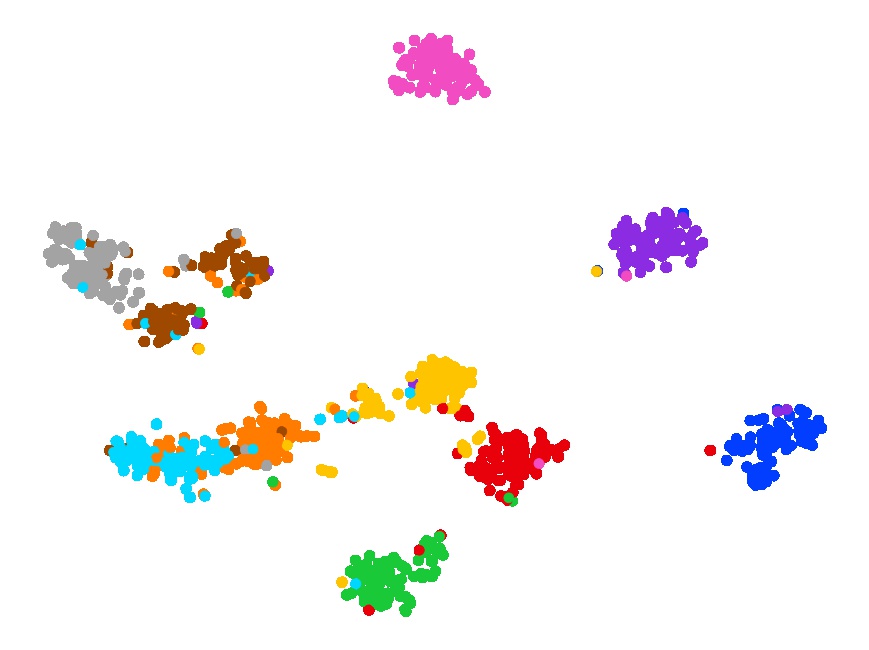}
        \caption{MTP-CLNN (Ours)}
        \label{subfigure: scatter_clnn3}
    \end{subfigure}
    \caption{t-SNE visulization of embeddings on M-CID. $\text{KCR}=25\%$, $\text{LAR}=10\%$. Best viewed in color.}
    \label{fig:scatter3}
\end{figure*}

The results on semi-supervised NID when $LAR=50\%$ are shown in Table~\ref{tab:main_semi2}. It can be seen that our methods still achieve the best performance in this case.
In Fig.~\ref{fig:scatter2} and Fig.~\ref{fig:scatter3}, we show the t-SNE visualization of clusters on BANKING and M-CID with embeddings learned by two strongest baselines and our methods. Again, it shows that our methods can produce more compact clusters.

\input{tables/main_semi2}

\input{tables/fig4}


%% file: tables/main_semi2.tex
\begin{table*}[ht!]
\centering
\scalebox{0.85}{
\begin{tabular}{cc|ccc|ccc|ccc}
\specialrule{0.1em}{0.2em}{0.2em}
\multicolumn{2}{c}{}&\multicolumn{3}{c}{BANKING}&\multicolumn{3}{c}{StackOverflow}&\multicolumn{3}{c}{M-CID}\\
KCR&Methods & NMI & ARI & ACC & NMI & ARI & ACC & NMI & ARI & ACC\\
\specialrule{0.05em}{0.1em}{0.2em}
\multirow{8}{*}{25\%} 
                    &BERT-DTC& 66.66&32.47&45.03 & 36.66&20.85&35.43 & 40.52&17.49&32.21   \\
                    &BERT-KCL& 56.88&19.40&23.28 & 35.26&16.18&30.86 & 31.81&11.13&23.18  \\
                    &BERT-MCL& 52.34&17.30&24.82 & 34.30&19.12&34.68 & 31.28&11.69&26.79 \\
                    &CDAC+& 71.71&39.60&53.25 & 73.92&39.29&72.76 & 32.22&12.98&30.49 \\
                    &DAC& 73.89&42.84&55.01 & 56.80&39.51&55.33 & 53.72&31.36&47.36 \\
                    & MTP \textbf{(Ours)} & 80.94&53.44&63.68 & 75.30&63.27&76.97 & 73.77&55.37&69.00 \\
                    & MTP-DAC \textbf{(Comb)} & 83.05&58.36&68.17 & 78.15&64.64&79.25 & 78.36&64.33&77.36 \\
                    & MTP-CLNN \textbf{(Ours)} & \bf85.30&\bf64.12&\bf73.76 & \bf80.15&\bf71.29&\bf84.56 & \bf80.61&\bf67.31&\bf79.34 \\
\specialrule{0.05em}{0.1em}{0.2em}
\multirow{8}{*}{50\%} 
                    &BERT-DTC& 76.32&48.04&60.35 & 55.76&39.46&55.58 & 57.62&35.66&50.40\\
                    &BERT-KCL& 70.60&37.47&45.37 & 55.82&37.29&48.32 & 53.88&32.02&44.76\\
                    &BERT-MCL& 69.52&37.04&45.65 & 53.75&38.58&52.04 & 48.66&28.38&45.67\\
                    &CDAC+& 76.15&47.01&59.31 & 74.87&39.38&74.38 & 51.58&28.96&48.08\\
                    &DAC& 79.89&54.09&65.14 & 72.51&59.12&74.51 & 68.21&49.59&63.84 \\
                    &MTP \textbf{(Ours)}& 85.03&62.97&72.63 & 79.58&70.49&83.51 & 78.53&63.93&76.50 \\
                    &MTP-DAC \textbf{(Comb)}& 86.78&67.23&76.48 & 81.36&72.58&85.18 & 81.42&69.36&81.98 \\
                    &MTP-CLNN \textbf{(Ours)}& \bf88.09&\bf71.07&\bf80.38 & \bf82.84&\bf75.54&\bf87.21 & \bf82.46&\bf71.21&\bf82.75 \\
\specialrule{0.05em}{0.1em}{0.2em}
\multirow{8}{*}{75\%}
                    &BERT-DTC& 81.60&58.00&69.47 & 69.61&57.10&72.56 & 69.57&51.19&66.39   \\
                    &BERT-KCL& 81.23&57.94&68.33 & 66.85&53.02&66.14 & 68.46&50.21&64.67  \\
                    &BERT-MCL& 80.98&57.72&68.46 & 68.39&54.40&65.98 & 64.09&45.82&63.21 \\
                    &CDAC+& 77.76&49.59&61.48 & 76.09&41.37&76.64 & 65.72&39.95&63.87 \\
                    &DAC& 84.78&64.25&74.09 & 77.67&65.45&81.45 & 77.37&63.84&77.05 \\
                    & MTP \textbf{(Ours)} & 88.69&72.65&81.24 & 83.27&76.19&87.12 & 83.53&73.45&83.87 \\
                    & MTP-DAC \textbf{(Comb)}& 89.52&74.57&82.97 & 83.97&77.24&87.91 & 84.89&75.84&86.59 \\
                    & MTP-CLNN \textbf{(Ours)} & \bf90.51&\bf77.55&\bf85.18 & \bf84.46&\bf78.39&\bf88.98 & \bf85.78&\bf77.40&\bf87.74 \\
\specialrule{0.1em}{0.1em}{0.2em}
\end{tabular}
}
\caption{\label{tab:main_semi2}
Performance on semi-supervised NID with different known class ratio. The $\text{LAR}$ is set to $50\%$. For each dataset, the best results are marked in bold. \textbf{Comb} denotes the baseline method combined with our proposed MTP.
}
\end{table*}

%% file: tables/fig4.tex
\begin{table*}[ht!]
\centering
\scalebox{0.85}{
\begin{tabular}{cc|ccc|ccc|ccc}
\specialrule{0.1em}{0.2em}{0.2em}
\multicolumn{2}{c}{}&\multicolumn{3}{c}{BANKING}&\multicolumn{3}{c}{StackOverflow}&\multicolumn{3}{c}{M-CID}\\
KCR&Methods & NMI & ARI & ACC & NMI & ARI & ACC & NMI & ARI & ACC\\
\specialrule{0.05em}{0.1em}{0.2em}
\multirow{2}{*}{0\%} 
                    &SUP-CLNN& 40.74 & 7.50 & 18.39 & 10.35 & 1.27 & 13.70 & 49.08 & 25.16 & 42.87 \\
                    &MTP-CLNN& 81.80 & 55.75 & 65.90 & 78.71 & 67.63 & 81.43 & 79.95 & 66.71 & 79.14 \\
\specialrule{0.05em}{0.1em}{0.2em}
\multirow{2}{*}{25\%} 
                    &SUP-CLNN& 80.91 & 55.15 & 66.05 & 75.65 & 58.23 & 77.22 & 47.56 & 23.42 & 41.23 \\
                    &MTP-CLNN& 84.11 & 61.29 & 71.43 & 79.68 & 70.17 & 83.77 & 80.24 & 66.77 & 79.20 \\
\specialrule{0.05em}{0.1em}{0.2em}
\multirow{2}{*}{50\%} 
                    &SUP-CLNN& 84.19 & 62.13 & 73.01 & 79.77 & 69.70 & 83.91 & 47.97 & 23.91 & 42.81 \\
                    &MTP-CLNN& 85.62 & 64.93 & 75.23 & 81.03 & 73.02 & 85.64 & 79.48 & 77.65 & 77.85 \\
\specialrule{0.05em}{0.1em}{0.2em}
\multirow{2}{*}{75\%} 
                    &SUP-CLNN& 86.56 & 68.15 & 78.10 & 81.61 & 74.35 & 86.94 & 75.55 & 60.56 & 73.24 \\
                    &MTP-CLNN& 87.52 & 70.00 & 79.74 & 82.56 & 75.66 & 87.63 & 83.75 & 73.22 & 84.36 \\
\specialrule{0.1em}{0.1em}{0.2em}
\end{tabular}
}
\caption{\label{tab:fig4}
Ablation study on the effectiveness of MTP. The LAR is set to 10\%. SUP stands for supervised pre-training
on internal labeled data only.
}
\end{table*}

%% file: acl.bbl
\begin{thebibliography}{53}
\expandafter\ifx\csname natexlab\endcsname\relax\def\natexlab#1{#1}\fi

\bibitem[{Arora et~al.(2020)Arora, Shrivastava, Mohit, Lecanda, and Aly}]{arora2020cross}
Abhinav Arora, Akshat Shrivastava, Mrinal Mohit, Lorena Sainz-Maza Lecanda, and Ahmed Aly. 2020.
\newblock Cross-lingual transfer learning for intent detection of covid-19 utterances.

\bibitem[{Bachman et~al.(2019)Bachman, Hjelm, and Buchwalter}]{bachman2019cpc}
Philip Bachman, R~Devon Hjelm, and William Buchwalter. 2019.
\newblock \href {https://proceedings.neurips.cc/paper/2019/file/ddf354219aac374f1d40b7e760ee5bb7-Paper.pdf} {Learning representations by maximizing mutual information across views}.
\newblock In \emph{Advances in Neural Information Processing Systems}, volume~32. Curran Associates, Inc.

\bibitem[{Brown et~al.(2020)Brown, Mann, Ryder, Subbiah, Kaplan, Dhariwal, Neelakantan, Shyam, Sastry, Askell, Agarwal, Herbert-Voss, Krueger, Henighan, Child, Ramesh, Ziegler, Wu, Winter, Hesse, Chen, Sigler, Litwin, Gray, Chess, Clark, Berner, McCandlish, Radford, Sutskever, and Amodei}]{brown2020gpt3}
Tom Brown, Benjamin Mann, Nick Ryder, Melanie Subbiah, Jared~D Kaplan, Prafulla Dhariwal, Arvind Neelakantan, Pranav Shyam, Girish Sastry, Amanda Askell, Sandhini Agarwal, Ariel Herbert-Voss, Gretchen Krueger, Tom Henighan, Rewon Child, Aditya Ramesh, Daniel Ziegler, Jeffrey Wu, Clemens Winter, Chris Hesse, Mark Chen, Eric Sigler, Mateusz Litwin, Scott Gray, Benjamin Chess, Jack Clark, Christopher Berner, Sam McCandlish, Alec Radford, Ilya Sutskever, and Dario Amodei. 2020.
\newblock \href {https://proceedings.neurips.cc/paper/2020/file/1457c0d6bfcb4967418bfb8ac142f64a-Paper.pdf} {Language models are few-shot learners}.
\newblock In \emph{Advances in Neural Information Processing Systems}, volume~33, pages 1877--1901. Curran Associates, Inc.

\bibitem[{Caron et~al.(2018)Caron, Bojanowski, Joulin, and Douze}]{caron2018deep}
Mathilde Caron, Piotr Bojanowski, Armand Joulin, and Matthijs Douze. 2018.
\newblock Deep clustering for unsupervised learning of visual features.
\newblock In \emph{Proceedings of the European Conference on Computer Vision (ECCV)}.

\bibitem[{Casanueva et~al.(2020)Casanueva, Tem{\v{c}}inas, Gerz, Henderson, and Vuli{\'c}}]{casanueva2020efficient}
I{\~n}igo Casanueva, Tadas Tem{\v{c}}inas, Daniela Gerz, Matthew Henderson, and Ivan Vuli{\'c}. 2020.
\newblock \href {https://doi.org/10.18653/v1/2020.nlp4convai-1.5} {Efficient intent detection with dual sentence encoders}.
\newblock In \emph{Proceedings of the 2nd Workshop on Natural Language Processing for Conversational AI}, pages 38--45, Online. Association for Computational Linguistics.

\bibitem[{Chatterjee and Sengupta(2020)}]{chatterjee2020mining}
Ajay Chatterjee and Shubhashis Sengupta. 2020.
\newblock \href {https://doi.org/10.18653/v1/2020.coling-main.366} {Intent mining from past conversations for conversational agent}.
\newblock In \emph{Proceedings of the 28th International Conference on Computational Linguistics}, pages 4140--4152, Barcelona, Spain (Online). International Committee on Computational Linguistics.

\bibitem[{Chen et~al.(2020)Chen, Kornblith, Norouzi, and Hinton}]{chen2020simple}
Ting Chen, Simon Kornblith, Mohammad Norouzi, and Geoffrey Hinton. 2020.
\newblock \href {http://arxiv.org/abs/2002.05709} {A simple framework for contrastive learning of visual representations}.

\bibitem[{Cheung and Li(2012)}]{cheung2012sequence}
Jackie Chi~Kit Cheung and Xiao Li. 2012.
\newblock Sequence clustering and labeling for unsupervised query intent discovery.
\newblock In \emph{Proceedings of the fifth ACM international conference on Web search and data mining}, pages 383--392.

\bibitem[{Devlin et~al.(2019)Devlin, Chang, Lee, and Toutanova}]{devlin2019bert}
Jacob Devlin, Ming-Wei Chang, Kenton Lee, and Kristina Toutanova. 2019.
\newblock \href {https://doi.org/10.18653/v1/N19-1423} {{BERT}: Pre-training of deep bidirectional transformers for language understanding}.
\newblock In \emph{Proceedings of the 2019 Conference of the North {A}merican Chapter of the Association for Computational Linguistics: Human Language Technologies, Volume 1 (Long and Short Papers)}, pages 4171--4186, Minneapolis, Minnesota. Association for Computational Linguistics.

\bibitem[{Forman et~al.(2015)Forman, Nachlieli, and Keshet}]{forman2015semi}
George Forman, Hila Nachlieli, and Renato Keshet. 2015.
\newblock Clustering by intent: A semi-supervised method to discover relevant clusters incrementally.
\newblock In \emph{Machine Learning and Knowledge Discovery in Databases}, pages 20--36, Cham. Springer International Publishing.

\bibitem[{Gao et~al.(2021)Gao, Yao, and Chen}]{gao2021simcse}
Tianyu Gao, Xingcheng Yao, and Danqi Chen. 2021.
\newblock \href {https://aclanthology.org/2021.emnlp-main.552} {{S}im{CSE}: Simple contrastive learning of sentence embeddings}.
\newblock In \emph{Proceedings of the 2021 Conference on Empirical Methods in Natural Language Processing}, pages 6894--6910, Online and Punta Cana, Dominican Republic. Association for Computational Linguistics.

\bibitem[{Giorgi et~al.(2021)Giorgi, Nitski, Wang, and Bader}]{giorgi2021declutr}
John Giorgi, Osvald Nitski, Bo~Wang, and Gary Bader. 2021.
\newblock \href {https://doi.org/10.18653/v1/2021.acl-long.72} {{D}e{CLUTR}: Deep contrastive learning for unsupervised textual representations}.
\newblock In \emph{Proceedings of the 59th Annual Meeting of the Association for Computational Linguistics and the 11th International Joint Conference on Natural Language Processing (Volume 1: Long Papers)}, pages 879--895, Online. Association for Computational Linguistics.

\bibitem[{Gowda(1984)}]{gowda1984ag}
K~Chidananda Gowda. 1984.
\newblock A feature reduction and unsupervised classification algorithm for multispectral data.
\newblock \emph{Pattern recognition}, 17(6):667--676.

\bibitem[{Gunel et~al.(2021)Gunel, Du, Conneau, and Stoyanov}]{gunel2021supervised}
Beliz Gunel, Jingfei Du, Alexis Conneau, and Veselin Stoyanov. 2021.
\newblock \href {https://openreview.net/forum?id=cu7IUiOhujH} {Supervised contrastive learning for pre-trained language model fine-tuning}.
\newblock In \emph{International Conference on Learning Representations}.

\bibitem[{Hakkani-T{\"u}r et~al.(2015)Hakkani-T{\"u}r, Ju, Zweig, and Tur}]{hakkani2015clustering}
Dilek Hakkani-T{\"u}r, Yun-Cheng Ju, Geoffrey Zweig, and Gokhan Tur. 2015.
\newblock Clustering novel intents in a conversational interaction system with semantic parsing.
\newblock In \emph{Sixteenth Annual Conference of the International Speech Communication Association}.

\bibitem[{Han et~al.(2019)Han, Vedaldi, and Zisserman}]{han2019dtc}
Kai Han, Andrea Vedaldi, and Andrew Zisserman. 2019.
\newblock Learning to discover novel visual categories via deep transfer clustering.
\newblock In \emph{Proceedings of the IEEE/CVF International Conference on Computer Vision (ICCV)}.

\bibitem[{Haponchyk and Moschitti(2021)}]{haponchyk2021structured}
Iryna Haponchyk and Alessandro Moschitti. 2021.
\newblock \href {https://doi.org/10.18653/v1/2021.naacl-main.263} {Supervised neural clustering via latent structured output learning: Application to question intents}.
\newblock In \emph{Proceedings of the 2021 Conference of the North American Chapter of the Association for Computational Linguistics: Human Language Technologies}, pages 3364--3374, Online. Association for Computational Linguistics.

\bibitem[{Haponchyk et~al.(2018)Haponchyk, Uva, Yu, Uryupina, and Moschitti}]{haponchyk2018structured}
Iryna Haponchyk, Antonio Uva, Seunghak Yu, Olga Uryupina, and Alessandro Moschitti. 2018.
\newblock \href {https://doi.org/10.18653/v1/D18-1254} {Supervised clustering of questions into intents for dialog system applications}.
\newblock In \emph{Proceedings of the 2018 Conference on Empirical Methods in Natural Language Processing}, pages 2310--2321, Brussels, Belgium. Association for Computational Linguistics.

\bibitem[{He et~al.(2019)He, Fan, Wu, Xie, and Girshick}]{he2019moco}
Kaiming He, Haoqi Fan, Yuxin Wu, Saining Xie, and Ross Girshick. 2019.
\newblock Momentum contrast for unsupervised visual representation learning.
\newblock \emph{arXiv preprint arXiv:1911.05722}.

\bibitem[{Henderson et~al.(2020)Henderson, Casanueva, Mrk{\v{s}}i{\'c}, Su, Wen, and Vuli{\'c}}]{henderson2020convert}
Matthew Henderson, I{\~n}igo Casanueva, Nikola Mrk{\v{s}}i{\'c}, Pei-Hao Su, Tsung-Hsien Wen, and Ivan Vuli{\'c}. 2020.
\newblock \href {https://doi.org/10.18653/v1/2020.findings-emnlp.196} {{C}onve{RT}: Efficient and accurate conversational representations from transformers}.
\newblock In \emph{Findings of the Association for Computational Linguistics: EMNLP 2020}, pages 2161--2174, Online. Association for Computational Linguistics.

\bibitem[{Hosseini{-}Asl et~al.(2020)Hosseini{-}Asl, McCann, Wu, Yavuz, and Socher}]{ehsan2020simpletod}
Ehsan Hosseini{-}Asl, Bryan McCann, Chien{-}Sheng Wu, Semih Yavuz, and Richard Socher. 2020.
\newblock \href {https://proceedings.neurips.cc/paper/2020/hash/e946209592563be0f01c844ab2170f0c-Abstract.html} {A simple language model for task-oriented dialogue}.
\newblock In \emph{Advances in Neural Information Processing Systems 33: Annual Conference on Neural Information Processing Systems 2020, NeurIPS 2020, December 6-12, 2020, virtual}.

\bibitem[{Hsu et~al.(2018)Hsu, Lv, and Kira}]{hsu2018kcl}
Yen-Chang Hsu, Zhaoyang Lv, and Zsolt Kira. 2018.
\newblock \href {https://openreview.net/forum?id=ByRWCqvT-} {Learning to cluster in order to transfer across domains and tasks}.
\newblock In \emph{International Conference on Learning Representations (ICLR)}.

\bibitem[{Hsu et~al.(2019)Hsu, Lv, Schlosser, Odom, and Kira}]{hsu2019mcl}
Yen-Chang Hsu, Zhaoyang Lv, Joel Schlosser, Phillip Odom, and Zsolt Kira. 2019.
\newblock \href {https://openreview.net/forum?id=SJzR2iRcK7} {Multi-class classification without multi-class labels}.
\newblock In \emph{International Conference on Learning Representations (ICLR)}.

\bibitem[{Johnson et~al.(2017)Johnson, Douze, and J{\'e}gou}]{johnson2017faiss}
Jeff Johnson, Matthijs Douze, and Herv{\'e} J{\'e}gou. 2017.
\newblock Billion-scale similarity search with gpus.
\newblock \emph{arXiv preprint arXiv:1702.08734}.

\bibitem[{Khosla et~al.(2020)Khosla, Teterwak, Wang, Sarna, Tian, Isola, Maschinot, Liu, and Krishnan}]{khosla2020supervised}
Prannay Khosla, Piotr Teterwak, Chen Wang, Aaron Sarna, Yonglong Tian, Phillip Isola, Aaron Maschinot, Ce~Liu, and Dilip Krishnan. 2020.
\newblock \href {https://proceedings.neurips.cc/paper/2020/file/d89a66c7c80a29b1bdbab0f2a1a94af8-Paper.pdf} {Supervised contrastive learning}.
\newblock In \emph{Advances in Neural Information Processing Systems}, volume~33, pages 18661--18673. Curran Associates, Inc.

\bibitem[{Kim et~al.(2021)Kim, Yoo, and Lee}]{kim2021guided}
Taeuk Kim, Kang~Min Yoo, and Sang-goo Lee. 2021.
\newblock \href {https://doi.org/10.18653/v1/2021.acl-long.197} {Self-guided contrastive learning for {BERT} sentence representations}.
\newblock In \emph{Proceedings of the 59th Annual Meeting of the Association for Computational Linguistics and the 11th International Joint Conference on Natural Language Processing (Volume 1: Long Papers)}, pages 2528--2540, Online. Association for Computational Linguistics.

\bibitem[{Larson et~al.(2019)Larson, Mahendran, Peper, Clarke, Lee, Hill, Kummerfeld, Leach, Laurenzano, Tang, and Mars}]{larson2019clinc}
Stefan Larson, Anish Mahendran, Joseph~J. Peper, Christopher Clarke, Andrew Lee, Parker Hill, Jonathan~K. Kummerfeld, Kevin Leach, Michael~A. Laurenzano, Lingjia Tang, and Jason Mars. 2019.
\newblock \href {https://doi.org/10.18653/v1/D19-1131} {An evaluation dataset for intent classification and out-of-scope prediction}.
\newblock In \emph{Proceedings of the 2019 Conference on Empirical Methods in Natural Language Processing and the 9th International Joint Conference on Natural Language Processing (EMNLP-IJCNLP)}, pages 1311--1316, Hong Kong, China. Association for Computational Linguistics.

\bibitem[{Lin et~al.(2020)Lin, Xu, and Zhang}]{lin2020constrained}
Ting-En Lin, Hua Xu, and Hanlei Zhang. 2020.
\newblock Discovering new intents via constrained deep adaptive clustering with cluster refinement.
\newblock In \emph{Thirty-Fourth AAAI Conference on Artificial Intelligence}.

\bibitem[{Liu et~al.(2019)Liu, Ott, Goyal, Du, Joshi, Chen, Levy, Lewis, Zettlemoyer, and Stoyanov}]{liu2019roberta}
Yinhan Liu, Myle Ott, Naman Goyal, Jingfei Du, Mandar Joshi, Danqi Chen, Omer Levy, Mike Lewis, Luke Zettlemoyer, and Veselin Stoyanov. 2019.
\newblock \href {http://arxiv.org/abs/1907.11692} {Roberta: A robustly optimized bert pretraining approach}.

\bibitem[{MacQueen et~al.(1967)}]{macqueen1967kmeans}
James MacQueen et~al. 1967.
\newblock Some methods for classification and analysis of multivariate observations.
\newblock In \emph{Proceedings of the fifth Berkeley symposium on mathematical statistics and probability}, volume~1, pages 281--297. Oakland, CA, USA.

\bibitem[{Mehri et~al.(2020)Mehri, Eric, and Hakkani-Tur}]{mehri2020dialoglue}
Shikib Mehri, Mihail Eric, and Dilek Hakkani-Tur. 2020.
\newblock \href {https://arxiv.org/abs/2009.13570} {Dialoglue: A natural language understanding benchmark for task-oriented dialogue}.
\newblock \emph{ArXiv preprint}, abs/2009.13570.

\bibitem[{Padmasundari(2018)}]{padmasundari2018intent}
Srinivas~Bangalore Padmasundari. 2018.
\newblock Intent discovery through unsupervised semantic text clustering.
\newblock \emph{Proc. Interspeech 2018}, pages 606--610.

\bibitem[{Pennington et~al.(2014)Pennington, Socher, and Manning}]{pennington2014glove}
Jeffrey Pennington, Richard Socher, and Christopher Manning. 2014.
\newblock \href {https://doi.org/10.3115/v1/D14-1162} {{G}lo{V}e: Global vectors for word representation}.
\newblock In \emph{Proceedings of the 2014 Conference on Empirical Methods in Natural Language Processing ({EMNLP})}, pages 1532--1543, Doha, Qatar. Association for Computational Linguistics.

\bibitem[{Perkins and Yang(2019)}]{perkins2019multiview}
Hugh Perkins and Yi~Yang. 2019.
\newblock \href {https://doi.org/10.18653/v1/D19-1413} {Dialog intent induction with deep multi-view clustering}.
\newblock In \emph{Proceedings of the 2019 Conference on Empirical Methods in Natural Language Processing and the 9th International Joint Conference on Natural Language Processing (EMNLP-IJCNLP)}, pages 4016--4025, Hong Kong, China. Association for Computational Linguistics.

\bibitem[{Radford and Narasimhan(2018)}]{radford2018gpt}
Alec Radford and Karthik Narasimhan. 2018.
\newblock Improving language understanding by generative pre-training.

\bibitem[{Rebuffi et~al.(2020)Rebuffi, Ehrhardt, Han, Vedaldi, and Zisserman}]{rebuffi2020lsdc}
Sylvestre-Alvise Rebuffi, Sebastien Ehrhardt, Kai Han, Andrea Vedaldi, and Andrew Zisserman. 2020.
\newblock Lsd-c: Linearly separable deep clusters.
\newblock \emph{arXiv}.

\bibitem[{Shi et~al.(2018)Shi, Chen, Sha, Li, Sun, Wang, and Zhang}]{shi2018auto}
Chen Shi, Qi~Chen, Lei Sha, Sujian Li, Xu~Sun, Houfeng Wang, and Lintao Zhang. 2018.
\newblock \href {https://doi.org/10.18653/v1/D18-1072} {Auto-dialabel: Labeling dialogue data with unsupervised learning}.
\newblock In \emph{Proceedings of the 2018 Conference on Empirical Methods in Natural Language Processing}, pages 684--689, Brussels, Belgium. Association for Computational Linguistics.

\bibitem[{Van~Gansbeke et~al.(2020)Van~Gansbeke, Vandenhende, Georgoulis, Proesmans, and Van~Gool}]{vangansbeke2020scan}
Wouter Van~Gansbeke, Simon Vandenhende, Stamatios Georgoulis, Marc Proesmans, and Luc Van~Gool. 2020.
\newblock Scan: Learning to classify images without labels.
\newblock In \emph{Proceedings of the European Conference on Computer Vision}.

\bibitem[{Vedula et~al.(2020)Vedula, Gupta, Alok, and Sridhar}]{vedula2020automatic}
Nikhita Vedula, Rahul Gupta, Aman Alok, and Mukund Sridhar. 2020.
\newblock \href {http://arxiv.org/abs/2006.01208} {Automatic discovery of novel intents \& domains from text utterances}.

\bibitem[{Vuli{\'c} et~al.(2021)Vuli{\'c}, Su, Coope, Gerz, Budzianowski, Casanueva, Mrk{\v{s}}i{\'c}, and Wen}]{vulic2021convfit}
Ivan Vuli{\'c}, Pei-Hao Su, Samuel Coope, Daniela Gerz, Pawe{\l} Budzianowski, I{\~n}igo Casanueva, Nikola Mrk{\v{s}}i{\'c}, and Tsung-Hsien Wen. 2021.
\newblock \href {https://aclanthology.org/2021.emnlp-main.88} {{ConvFiT:} {C}onversational fine-tuning of pretrained language models}.
\newblock In \emph{Proceedings of the 2021 Conference on Empirical Methods in Natural Language Processing}, pages 1151--1168, Online and Punta Cana, Dominican Republic. Association for Computational Linguistics.

\bibitem[{Wei and Zou(2019)}]{wei2019eda}
Jason Wei and Kai Zou. 2019.
\newblock \href {https://doi.org/10.18653/v1/D19-1670} {{EDA}: Easy data augmentation techniques for boosting performance on text classification tasks}.
\newblock In \emph{Proceedings of the 2019 Conference on Empirical Methods in Natural Language Processing and the 9th International Joint Conference on Natural Language Processing (EMNLP-IJCNLP)}, pages 6382--6388, Hong Kong, China. Association for Computational Linguistics.

\bibitem[{Wolf et~al.(2019)Wolf, Debut, Sanh, Chaumond, Delangue, Moi, Cistac, Rault, Louf, Funtowicz, Davison, Shleifer, von Platen, Ma, Jernite, Plu, Xu, Scao, Gugger, Drame, Lhoest, and Rush}]{wolf2019huggingfaces}
Thomas Wolf, Lysandre Debut, Victor Sanh, Julien Chaumond, Clement Delangue, Anthony Moi, Pierric Cistac, Tim Rault, Rémi Louf, Morgan Funtowicz, Joe Davison, Sam Shleifer, Patrick von Platen, Clara Ma, Yacine Jernite, Julien Plu, Canwen Xu, Teven~Le Scao, Sylvain Gugger, Mariama Drame, Quentin Lhoest, and Alexander~M. Rush. 2019.
\newblock \href {http://arxiv.org/abs/1910.03771} {Huggingface's transformers: State-of-the-art natural language processing}.

\bibitem[{Wu et~al.(2020)Wu, Hoi, Socher, and Xiong}]{wu2020tod}
Chien-Sheng Wu, Steven~C.H. Hoi, Richard Socher, and Caiming Xiong. 2020.
\newblock \href {https://doi.org/10.18653/v1/2020.emnlp-main.66} {{TOD}-{BERT}: Pre-trained natural language understanding for task-oriented dialogue}.
\newblock In \emph{Proceedings of the 2020 Conference on Empirical Methods in Natural Language Processing (EMNLP)}, pages 917--929, Online. Association for Computational Linguistics.

\bibitem[{Xie et~al.(2016)Xie, Girshick, and Farhadi}]{xie2016dec}
Junyuan Xie, Ross Girshick, and Ali Farhadi. 2016.
\newblock \href {https://proceedings.mlr.press/v48/xieb16.html} {Unsupervised deep embedding for clustering analysis}.
\newblock In \emph{Proceedings of The 33rd International Conference on Machine Learning}, volume~48 of \emph{Proceedings of Machine Learning Research}, pages 478--487, New York, New York, USA. PMLR.

\bibitem[{Xu et~al.(2015)Xu, Wang, Tian, Xu, Zhao, Wang, and Hao}]{xu2015short}
Jiaming Xu, Peng Wang, Guanhua Tian, Bo~Xu, Jun Zhao, Fangyuan Wang, and Hongwei Hao. 2015.
\newblock \href {https://doi.org/10.3115/v1/W15-1509} {Short text clustering via convolutional neural networks}.
\newblock In \emph{Proceedings of the 1st Workshop on Vector Space Modeling for Natural Language Processing}, pages 62--69, Denver, Colorado. Association for Computational Linguistics.

\bibitem[{Yan et~al.(2021)Yan, Li, Wang, Zhang, Wu, and Xu}]{yan2021consert}
Yuanmeng Yan, Rumei Li, Sirui Wang, Fuzheng Zhang, Wei Wu, and Weiran Xu. 2021.
\newblock \href {https://doi.org/10.18653/v1/2021.acl-long.393} {{C}on{SERT}: A contrastive framework for self-supervised sentence representation transfer}.
\newblock In \emph{Proceedings of the 59th Annual Meeting of the Association for Computational Linguistics and the 11th International Joint Conference on Natural Language Processing (Volume 1: Long Papers)}, pages 5065--5075, Online. Association for Computational Linguistics.

\bibitem[{Yang et~al.(2017)Yang, Fu, Sidiropoulos, and Hong}]{bo2017dcn}
Bo~Yang, Xiao Fu, Nicholas~D. Sidiropoulos, and Mingyi Hong. 2017.
\newblock \href {https://proceedings.mlr.press/v70/yang17b.html} {Towards k-means-friendly spaces: Simultaneous deep learning and clustering}.
\newblock In \emph{Proceedings of the 34th International Conference on Machine Learning}, volume~70 of \emph{Proceedings of Machine Learning Research}, pages 3861--3870. PMLR.

\bibitem[{Zhang et~al.(2021{\natexlab{a}})Zhang, Nan, Wei, Li, Zhu, McKeown, Nallapati, Arnold, and Xiang}]{zhang2021supporting}
Dejiao Zhang, Feng Nan, Xiaokai Wei, Shang-Wen Li, Henghui Zhu, Kathleen McKeown, Ramesh Nallapati, Andrew~O. Arnold, and Bing Xiang. 2021{\natexlab{a}}.
\newblock \href {https://doi.org/10.18653/v1/2021.naacl-main.427} {Supporting clustering with contrastive learning}.
\newblock In \emph{Proceedings of the 2021 Conference of the North American Chapter of the Association for Computational Linguistics: Human Language Technologies}, pages 5419--5430, Online. Association for Computational Linguistics.

\bibitem[{Zhang et~al.(2021{\natexlab{b}})Zhang, Li, Xu, Zhang, Zhao, and Gao}]{zhang2021textoir}
Hanlei Zhang, Xiaoteng Li, Hua Xu, Panpan Zhang, Kang Zhao, and Kai Gao. 2021{\natexlab{b}}.
\newblock \href {https://doi.org/10.18653/v1/2021.acl-demo.20} {{TEXTOIR}: An integrated and visualized platform for text open intent recognition}.
\newblock In \emph{Proceedings of the 59th Annual Meeting of the Association for Computational Linguistics and the 11th International Joint Conference on Natural Language Processing: System Demonstrations}, pages 167--174, Online. Association for Computational Linguistics.

\bibitem[{Zhang et~al.(2021{\natexlab{c}})Zhang, Xu, Lin, and Lyu}]{zhang2021aligned}
Hanlei Zhang, Hua Xu, Ting-En Lin, and Rui Lyu. 2021{\natexlab{c}}.
\newblock Discovering new intents with deep aligned clustering.
\newblock \emph{Proceedings of the AAAI Conference on Artificial Intelligence}, 35(16):14365--14373.

\bibitem[{Zhang et~al.(2021{\natexlab{d}})Zhang, Zhang, Zhan, Chen, Shi, Wu, and Lam}]{zhang2021effectiveness}
Haode Zhang, Yuwei Zhang, Li-Ming Zhan, Jiaxin Chen, Guangyuan Shi, Xiao-Ming Wu, and Albert Y.~S. Lam. 2021{\natexlab{d}}.
\newblock \href {http://arxiv.org/abs/2109.05782} {Effectiveness of pre-training for few-shot intent classification}.

\bibitem[{Zhang et~al.(2021{\natexlab{e}})Zhang, Bui, Yoon, Chen, Liu, Xia, Tran, Chang, and Yu}]{zhang2021contrastive}
Jianguo Zhang, Trung Bui, Seunghyun Yoon, Xiang Chen, Zhiwei Liu, Congying Xia, Quan~Hung Tran, Walter Chang, and Philip Yu. 2021{\natexlab{e}}.
\newblock \href {https://aclanthology.org/2021.emnlp-main.144} {Few-shot intent detection via contrastive pre-training and fine-tuning}.
\newblock In \emph{Proceedings of the 2021 Conference on Empirical Methods in Natural Language Processing}, pages 1906--1912, Online and Punta Cana, Dominican Republic. Association for Computational Linguistics.

\bibitem[{Zhang et~al.(2020)Zhang, Hashimoto, Liu, Wu, Wan, Yu, Socher, and Xiong}]{zhang2020discriminative}
Jianguo Zhang, Kazuma Hashimoto, Wenhao Liu, Chien-Sheng Wu, Yao Wan, Philip Yu, Richard Socher, and Caiming Xiong. 2020.
\newblock \href {https://doi.org/10.18653/v1/2020.emnlp-main.411} {Discriminative nearest neighbor few-shot intent detection by transferring natural language inference}.
\newblock In \emph{Proceedings of the 2020 Conference on Empirical Methods in Natural Language Processing (EMNLP)}, pages 5064--5082, Online. Association for Computational Linguistics.

\end{thebibliography}
